\documentclass{article} % For LaTeX2e
\usepackage{iclr2025_conference,times}

\usepackage{amsmath,amsfonts,bm}

\def\eqref#1{equation~\ref{#1}}
\def\1{\bm{1}}

\DeclareMathAlphabet{\mathsfit}{\encodingdefault}{\sfdefault}{m}{sl}
\SetMathAlphabet{\mathsfit}{bold}{\encodingdefault}{\sfdefault}{bx}{n}

\usepackage{xcolor} 

\definecolor{mydarkblue}{rgb}{0,0.08,0.45} 
\definecolor{myblue}{RGB}{90, 140, 210} 

\usepackage[colorlinks,
            linkcolor=myblue,    
            anchorcolor=blue, 
            citecolor=myblue, 
            urlcolor=magenta,
            ]{hyperref}
\usepackage{url}
\usepackage{amsmath,amssymb}
\usepackage{amsthm} 
\usepackage{subcaption}
\usepackage{tabularx}
\usepackage{booktabs}
\usepackage{graphicx}
\usepackage{colortbl}
\usepackage{float}
\usepackage{algorithm}
\usepackage{multirow}
\usepackage{algpseudocode}
\usepackage[most]{tcolorbox}
\definecolor{commentblue}{RGB}{63, 134, 211}
\definecolor{algblue}{RGB}{236, 244, 255}    
\definecolor{tablegray}{RGB}{211, 211 , 211}
\definecolor{toyblue}{HTML}{3F86D3}
\definecolor{toybluebg}{HTML}{F4F6F8}
\newcommand{\hlbig}[1]{\cellcolor{commentblue}\textcolor{white}{#1}} 
\newcommand{\hlsmall}[1]{\cellcolor{algblue}{#1}}                  % 
\newcommand{\hlgray}[1]{\cellcolor{tablegray}{#1}}
\newtcolorbox{toyexamplebox}[2][]{
    enhanced,
    breakable,
    colback=toybluebg,
    colframe=toyblue,
    colbacktitle=toyblue,
    coltitle=white,
    fonttitle=\bfseries,
    title={#2},
    boxrule=0.8pt,
    arc=1.5mm,
    left=2mm,
    right=2mm,
    top=1.5mm,
    bottom=1.5mm,
    before skip=6pt,
    after skip=6pt,
    #1
}

\title{Critic-free Pretraining for Efficient Online Reinforcement Learning Fine-tuning}

\author{
Daoyi Li\textsuperscript{*} \quad
Yixian Zhang\textsuperscript{*} \quad
Yu Wang \textsuperscript{} \quad
Wenbo Ding\textsuperscript{ }\quad
Chao Yu\textsuperscript{\ensuremath{\dagger}}
\\
Tsinghua University
\\[2mm]
\textsuperscript{*}\textit{Equal contribution.}\qquad
\textsuperscript{\ensuremath{\dagger}}\textit{Corresponding authors.}
}

\iclrfinalcopy 
\begin{document}
\maketitle
\fancyhead{}
\fancyhead[L]{Preprint}
\begin{figure}[H]
    \centering
    \includegraphics[
        width=\linewidth
    ]{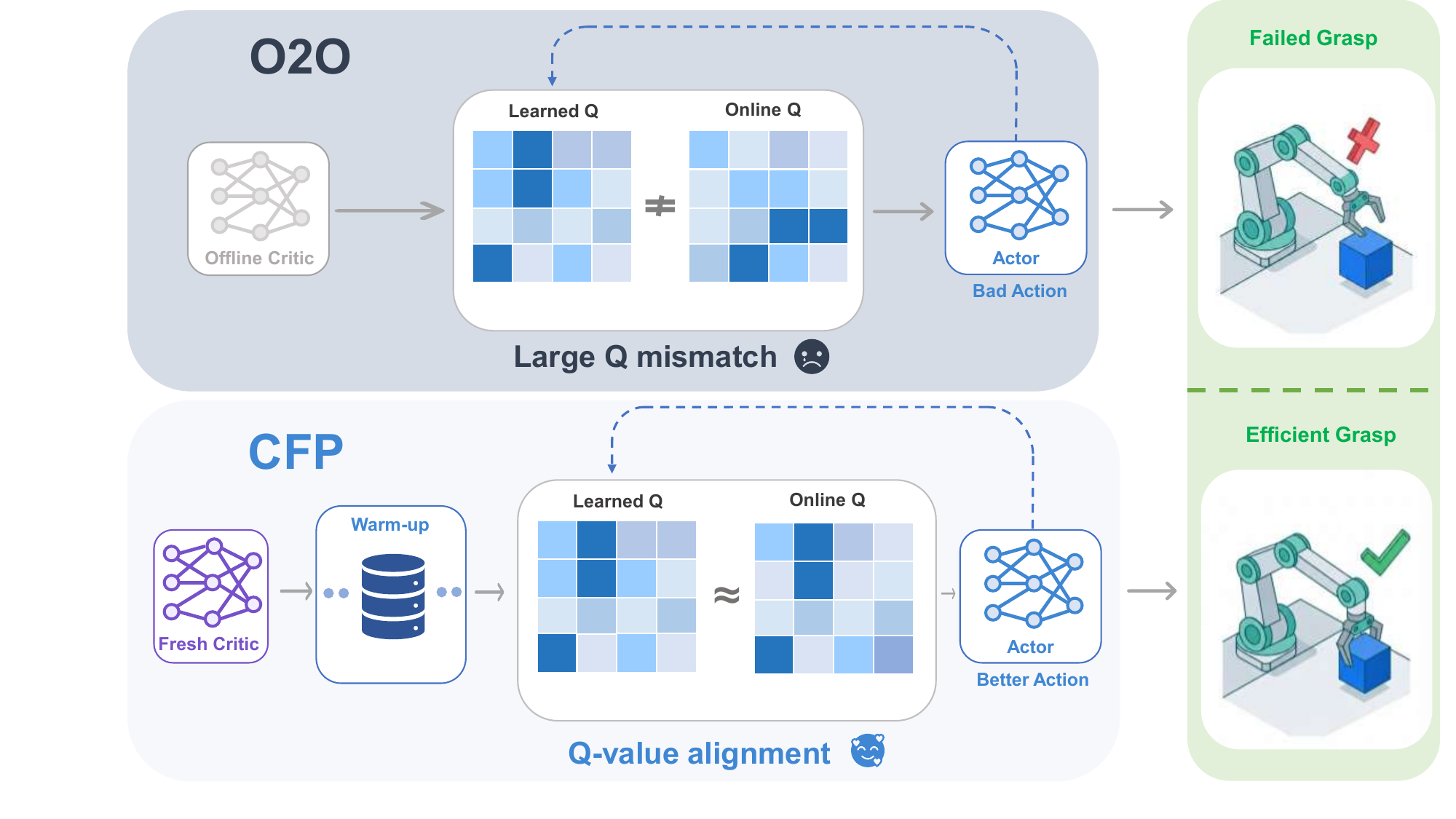}
    \caption{
        Overview of conventional offline-to-online fine-tuning and CFP. Conventional O2O directly reuses the offline-trained critic, which can become misaligned with the rapidly changing online policy. CFP instead retains the pretrained actor, restores a fresh critic, and calibrates it using a short warm-up before fine-tuning.
    }
    \label{fig:overview}
\end{figure}

\begin{abstract}
Offline-to-online (O2O) reinforcement learning aims to leverage policies pretrained on static datasets while improving them through online interaction. However, directly reusing an offline-trained critic can hinder online fine-tuning: as the policy and data distribution change rapidly, value estimates inherited from offline training may become misaligned with the online environment, leading to inaccurate policy improvement and inefficient exploration. To address this problem, we introduce \textbf{C}ritic-\textbf{F}ree \textbf{P}retraining: an efficient paradigm that completely abandons the approach of offline critic training, allowing a freshly initialized critic to adapt without inheriting biased estimates. CFP is compatible with various mainstream O2O algorithms and consistently matches or improves upon conventional O2O algorithms across a diverse set of tasks, with particularly pronounced gains on several challenging tasks.
\end{abstract}

\section{Introduction}

Reinforcement learning (RL) provides a general framework for learning
decision-making policies through interactions with an environment \citep{sutton2018reinforcement,haarnoja2018soft,DBLP:journals/corr/abs-1812-05905,kaelbling1996reinforcement,li2017deep,arulkumaran2017deep}. However, learning from scratch remains highly sample-inefficient in complex environments. At the beginning of training, an online agent rarely visits task-relevant regions. The agent must therefore improve its policy using low-quality experience, while the quality of future experience itself depends on the current policy. This coupling
between policy learning and data collection creates a fundamental problem: without a meaningful policy, the agent cannot efficiently collect useful data, and without useful data, it cannot learn a
meaningful policy.

The introduction of an offline phase significantly alleviates the pressure of exploration \citep{levine2020offline}. Offline training on a dataset greatly enhances the agent's training efficiency during the online phase \citep{fujimoto2021minimalist}. However, the O2O paradigm faces a critical issue: the transition from the dataset to the real-world environment often encounters significant discrepancies in data distribution, whereas the inherited critic reflects the offline behavior and data support. Its value estimates may consequently become misaligned with the evolving online policy, leading to inefficient policy updates, and even degradation of the
useful behavior acquired during offline pretraining. This represents a major challenge for O2O approaches \citep{kumar2020conservative,nakamoto2023cal,kostrikovoffline}.
 
In this work, we revisit the role of offline stage in O2O RL. Our insight is that the goal of offline training is not to produce a generalizable policy or value network, but rather to encourage the agent to operate in regions where it is more likely to collect effective trajectories, thereby improving sampling efficiency. This suggests that policy pretraining and critic pretraining, which are conventionally performed together, should be decoupled. Based on this insight, we propose \textbf{Critic-Free Pretraining} (CFP). 
Drawing inspiration from the concept of imitation learning, we utilize the dataset to train a behavior-cloning (BC) actor. Before online interaction begins, CFP retains the pretrained actor and introduces a fresh critic. A short warm-up phase then calibrates the critic. The pretrained policy provides efficient data collection, while the fresh critic avoids inheriting value bias tied to the offline distribution.

Our main contributions are listed below:
\begin{itemize}
    \item \textbf{A critic-free offline training paradigm.} 
    We demonstrate that training the critic during the offline phase can actually impair subsequent online fine-tuning performance. By completely eliminating offline critic training, we substantially reduce computational costs and memory requirements, improving training efficiency while providing novel insights for scaling large-scale reinforcement learning systems.

    \item \textbf{An effective yet simple implementation of CFP.} 
    We conduct extensive experiments on the implementation of CFP and identify an execution approach that balances computational efficiency with experimental performance. Building on this, we designed a toy example to simulate the training workflows of O2O and CFP, intuitively demonstrating the effectiveness of CFP through visualization. This work also offers a new methodology for future research into efficient reinforcement learning.

    \item \textbf{Strong and broadly applicable online fine-tuning performance.} 
    CFP achieves comparable and outstanding performance across multiple O2O algorithms and diverse benchmarks. The superior performance across different algorithms and task domains validates the effectiveness and generality of our approach.
\end{itemize}

\section{Preliminaries}
\label{Preliminaries}
\subsection{Reinforcement Learning}
We consider reinforcement learning in a Markov Decision Process (MDP) defined by the tuple $\mathcal{M}=(\mathcal{S},\mathcal{A},P,r,\gamma,\rho)$. Here, $\mathcal{S}$ and $\mathcal{A}$ denote the state and action spaces; $P(s'|s, a)$ represents the transition dynamics; $r(s, a)$ is the reward function; $\gamma \in [0, 1)$ is the discount factor; and $\rho$ is the initial state distribution. RL aims to find an policy $\pi^*(a|s)$ that maximizes the expected discounted reward:
\begin{equation}
J(\pi) = \mathbb{E}_{s_0 \sim \rho, a_t \sim \pi(\cdot|s_t), s_{t+1} \sim P(\cdot|s_t,a_t)} \left[ \sum_{t=0}^\infty \gamma^t r(s_t, a_t) \right].
\end{equation}

\subsection{Flow matching}
Flow Matching is a simple framework to train continuous normalizing flows (CNFs) \citep{lipman2023flow,lipman2024flow,liu2023flow,mcallister2026flow,yang2026policyflow}. In contrast to denoising diffusion models \citep{wang2023diffusion, ren2025diffusion, chi2025diffusion, kang2023efficient, wagenmaker2025steering}, which aim at solving Stochastic Differential Equations (SDEs), Flow Matching model learns a time-dependent vector field $v_\theta(t,x):[0,1]\times \mathbb{R}^d\to \mathbb{R}^d$, where $d$ is the distribution dimension, that characterizes the velocity of points moving from a noise distribution to the data distribution \citep{fql_park2025}. Generation is then performed by solving the ODE:
\begin{equation}
\frac{d x_t}{dt}
=
v_\theta(t,x_t),
\qquad
x_0\sim\rho_0.
\end{equation}
\subsection{Flow-based policies}
Flow-based policy train the network via Flow Matching. The objective is to  minimize the loss function:
\begin{equation}
L_{Flow}(\theta)=\mathbb{E}_{\substack{
    x_0 \sim \mathcal{N}(0, I_d), \\ 
    (x_1=a, s) \sim \mathcal{D}, \\ 
    t \sim \text{Unif}([0,1])
}}[\| v_\theta(t, s,x_t) - (x_1 - x_0) \|_2^2]
\end{equation}
where $\mathcal{D}$ denotes the static dataset. In inference, policies generate the action depend on the learned velocity field by computing the flow:
\begin{equation}
x_{t_{i+1}}
=
x_{t_i}
+
(t_{i+1}-t_i)
v_\theta(t_i,s,x_{t_i}),
\qquad
x_{t_0}\sim\mathcal N(0,I).
\end{equation}
In actual implementation, generating each action requires K-times computations of the vector field, which can lead to vanishing or exploding gradient issues during neural network training. Consequently, single-step policies $\mu_w(s,z)$ have emerged as a crucial solution. Rather than learning the velocity field directly, a single-step network learns to map to the final actions produced by $K$-step network. This allows actions that previously required multiple computations to be obtained in a single pass, significantly enhancing efficiency and enhancing training stability \citep{fql_park2025,zhang2026reinflow,zhang2026sac,li2025dqc, li2026qlearning,li2026reinforcement,doo2026qflow}. The overall loss function of flow-based policies is as below:
\begin{equation}
L_{Total}(\theta, \omega)=L_{Flow}(\theta)+L_{\pi}(\omega)
\end{equation}
where
\begin{equation}
L_{\pi}(\omega)=\alpha\mathbb{E}_{s\sim \mathcal{D},z\sim \mathcal{N}(0,I^d)}[\|\mu_\omega(s,z)-\mu_\theta(s,z)\|_2^2]+\mathbb{E}_{s\sim \mathcal{D},a\sim \pi_{\omega}(\cdot|s)}[-Q_\phi(s,a)] 
\label{eq:actor_loss}
\end{equation}

\section{Efficient online fine-tuning via CFP}
We present our main methodology in this section. Our method entirely omits critic training during the offline phase, resulting
in a critic-free policy-pretraining strategy. At the beginning of online fine-tuning, we initialize a fresh critic from scratch and perform a short warm-up stage on the offline dataset. 

\subsection{Critic-Free Pretraining}
Unlike previous O2O methods that update both the Actor and Critic simultaneously in the offline phase, we forgo training the latter in this phase. This also means that the actor's update in the offline phase does not receive gradients provided by the critic; that is, the actor's update formula is:

\begin{equation}
L_{\pi}(\omega)=\alpha\mathbb{E}_{s\sim \mathcal{D},z\sim \mathcal{N}(0,I^d)}[\|\mu_\omega(s,z)-\mu_\theta(s,z)\|_2^2]
\label{eq:actor_loss_bc}
\end{equation}
we reach a BC actor loss function in flow policies. BC is a widely-used algorithm that reaches outstanding performance in O2O situations and robot manipulation \citep{bhatt2026rainbow,torabi2018behavioral,foster2024behavior,sasaki2021behavioral}.

\subsection{
Why a Fresh Critic facilitates online fine-tuning}
\label{sec: why initialized critic helps with online Fine-tuning}
The ability of the critic to accurately describe the true Q-values is considered key to the algorithm's success. We offer a deeper insight: since an offline critic can hinder an agent's online fine-tuning capabilities, we can achieve our objective by simply initializing the critic, rather than expending effort on calibrating it. The rationale behind this idea is that a freshly initialized critic is initially inaccurate, but it does not inherit the systematic value bias induced by prolonged fitting to the offline distribution. However, such a fresh critic may fail to provide proper guidance to the actor and could even disrupt the pretrained offline policy, leading to instability during online training. To address this, we introduce a brief warm-up phase in which the critic is trained on offline data to acquire a preliminary ability to estimate Q-values. We present the comparison between O2O and CFP in Alg.\ref{alg:trad_o2o} and Alg.\ref{alg:our_cfp}.

\begin{figure*}[ht]
\centering

\begin{minipage}[t]{0.48\textwidth}
\null
\begin{algorithm}[H]
\caption{Traditional O2O}
\label{alg:trad_o2o}
\begin{algorithmic}[1]
\State \textbf{Init:} Critic $Q_\phi$, Policy $\pi_\theta$, Buffer $\mathcal{B} \leftarrow \mathcal{D}$

\Statex \textcolor{commentblue}{$\triangleright$ Offline Stage}
\For{$\ell = 1 \dots L_{\text{off}}$}
    \State Sample batch $\{(s, a, r, s')\} \sim \mathcal{B}$
    \State Update $\pi_\theta$ via offline actor loss 
    \State Update $Q_\phi$ via TD loss
\EndFor

\Statex \textcolor{commentblue}{$\triangleright$ Online Stage}
\For{$\ell = 1 \dots L_{\text{on}}$}
    \State Collect $(s, a, r, s') \sim \pi_\theta$, append to $\mathcal{B}$
    \State Sample batch $\{(s, a, r, s')\} \sim \mathcal{B}$
    \State Update $\pi_\theta$ via online actor loss
    \State Update $Q_\phi$ via TD loss
\EndFor
\end{algorithmic}
\end{algorithm}
\end{minipage}
\hfill
\begin{minipage}[t]{0.48\textwidth}
\null
\begin{algorithm}[H]
\caption{CFP (Ours)}
\label{alg:our_cfp}
\begin{algorithmic}[1]
\State \textbf{Init:} Policy $\pi_\theta$, Buffer $\mathcal{B} \leftarrow \mathcal{D}$ 
\Statex \textcolor{commentblue}{$\triangleright$ Offline Stage (Critic-Free)}
\For{$\ell = 1 \dots L_{\text{off}}$}
    \State Sample batch $\{(s, a, r, s')\} \sim \mathcal{B}$
    \State Update $\pi_\theta$ via BC Flow loss \hfill (Eq. \ref{eq:actor_loss_bc})
\EndFor
\Statex \colorbox{algblue}{%
\parbox{\dimexpr\linewidth-2\fboxsep}{%
    \textcolor{commentblue}{$\triangleright$ \textbf{Warm-up Stage}} \\
    \textbf{Init} fresh critic $Q_\phi$ from scratch \\
    \textbf{for} $\ell = 1 \dots L_{\text{warm-up}}$ \textbf{do} \\
    \phantom{aa} Sample batch $\{(s, a, r, s')\} \sim \mathcal{B}$ \\
    \phantom{aa} Update $\pi_\theta$ via online actor loss \\
    \phantom{aa} Update $Q_\phi$ via TD loss 
}}

\Statex \textcolor{commentblue}{$\triangleright$ Online Stage}
\For{$\ell = 1 \dots L_{\text{on}}$}
    \State Collect $(s, a, r, s') \sim \pi_\theta$, append to $\mathcal{B}$
    \State Sample batch $\{(s, a, r, s')\} \sim \mathcal{B}$
    \State Update $\pi_\theta$ via online actor loss
    \State Update $Q_\phi$ via TD loss
\EndFor
\end{algorithmic}
\end{algorithm}
\end{minipage}
\end{figure*}

Even though flow algorithms do not contain explicit penalties in loss functions, we assume that the inherent limitations of offline training are the implicit cause of the critic's pessimism. Since:
\begin{equation}
Q_\beta(s,a) \leq Q^*(s,a)
\end{equation}
where $Q^*(s,a)$ is the optimal Q-value in the real environment. However, the problem extends far beyond this natural bound. In an offline dataset generated by behavior policies, the learned value function $Q_\beta$ is theoretically bounded by the sub-optimality of the dataset. Modern offline datasets provide complex and sparse-reward tasks to test the effectiveness of algorithms. These datasets are overwhelmingly dominated by suboptimal, random, or failed trajectories, with high-quality successful demonstrations being exceptionally rare \citep{ogbench_park2025}. Due to the bootstrapping nature of Temporal Difference learning, the low returns from these abundant suboptimal trajectories propagate backward throughout the state-action space, results in accumulated pessimism of the critic.

\begin{figure}[htbp]
    \centering
    \includegraphics[
        width=0.65\linewidth
    ]{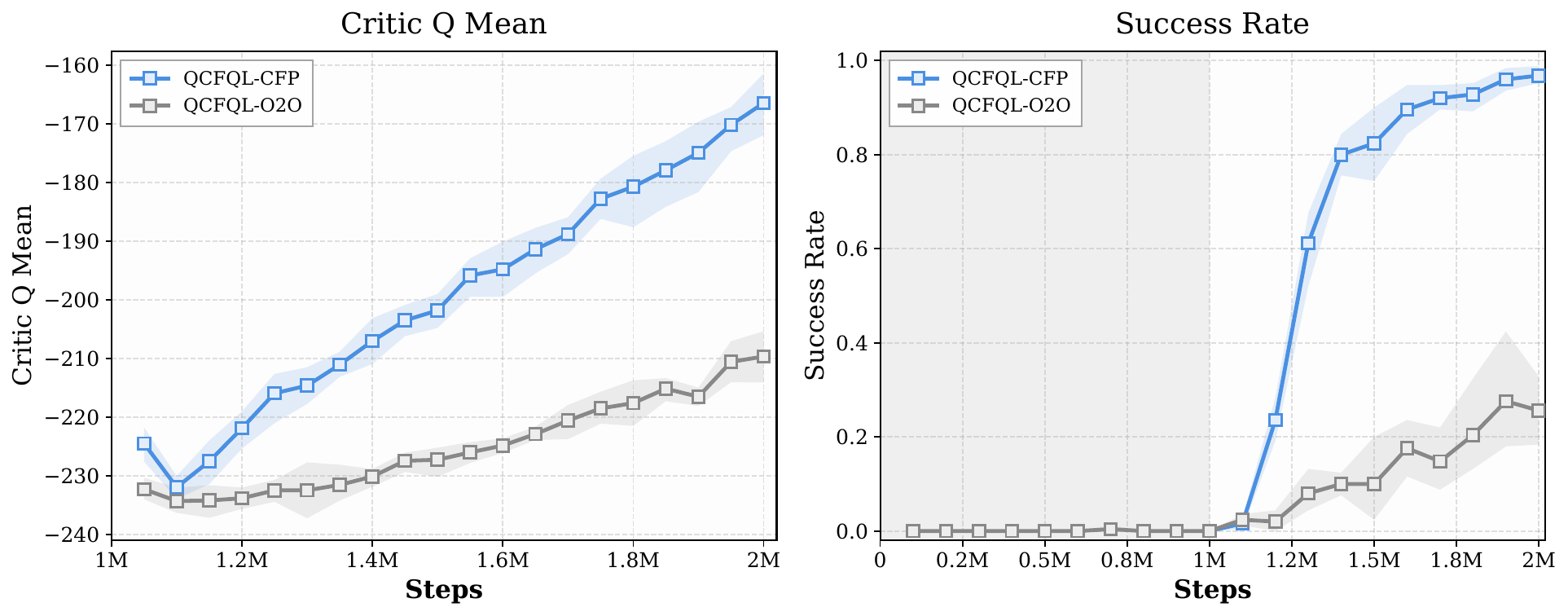}
    \caption{
        Critic Q-mean estimates and success rates on Cube Triple task 4. Curves and shaded regions show the mean and 95\% confidence interval across five seeds, respectively.
    }
    \label{fig:q-mean}
\end{figure}

As illustrated in Figure \ref{fig:q-mean}, we compared the Q-values calculated by the critic for state-action pairs in a batch under O2O and CFP. We find that during the online phase of CFP, the critic's Q-value improves rapidly as fine-tuning progressed; in contrast, the Q-value in the O2O consistently remains at a lower level.

Furthermore, we identified a potential mismatch between the critic's Q evaluations and the true Q values; this discrepancy prevents the critic from effectively guiding the actor toward superior actions. Given the complexity of the benchmark environments, we could not demonstrate this deviation through visualization; instead, we designed a toy example to illustrate our insight. As shown in Box \ref{box:toy-example}, we construct a tabular MDP with 16 states arranged on a $4\times4$ grid. The environment contains a negative terminal state, a positive terminal state, and several hazardous states. We deliberately induce a mismatch between the dataset Q distribution and the ground-truth Q distribution. We then separately simulate the training processes of O2O and CFP and evaluate the training performance of their respective critics.

\begin{toyexamplebox}[label={box:toy-example}]{Toy Example: Critic Mismatch under Distribution Shift}

\centering
\includegraphics[
    width=0.9\linewidth
]{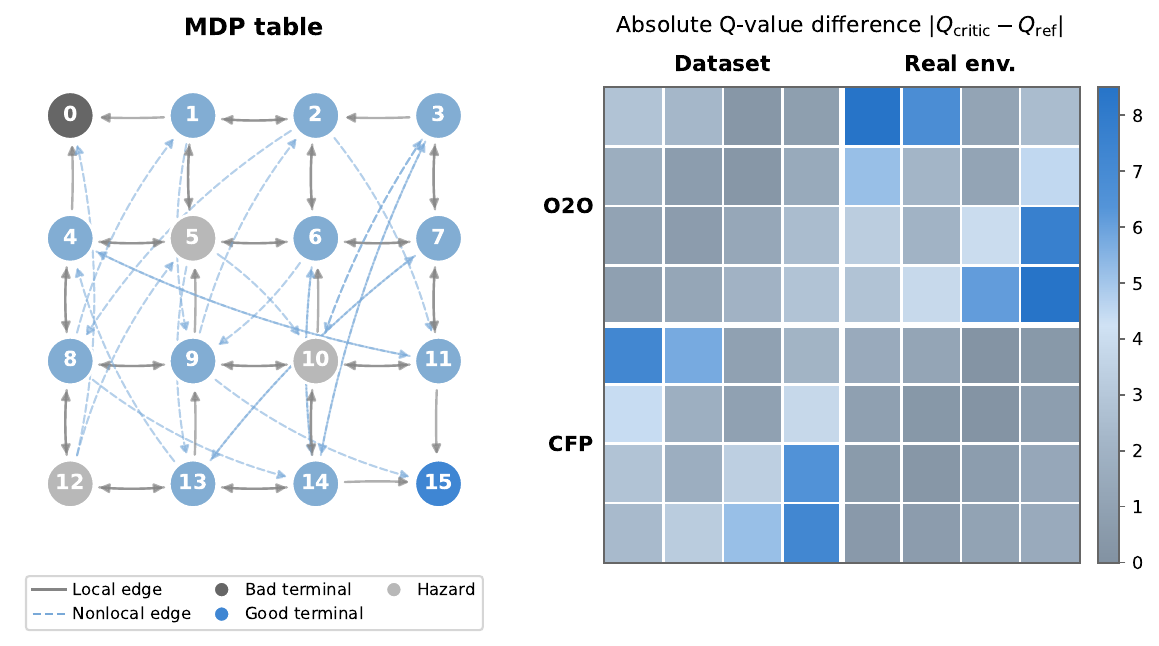}

\vspace{1mm}
\raggedright
\small
Schematic diagrams illustrating the MDP and Q-mismatch. The left panel represents the underlying transition dynamics within the MDP process. In the right panel, colors closer to blue indicate a greater degree of mismatch.
\end{toyexamplebox}

\begin{figure}[htbp]
    \centering
    \includegraphics[
        width=0.6\textwidth 
    ]{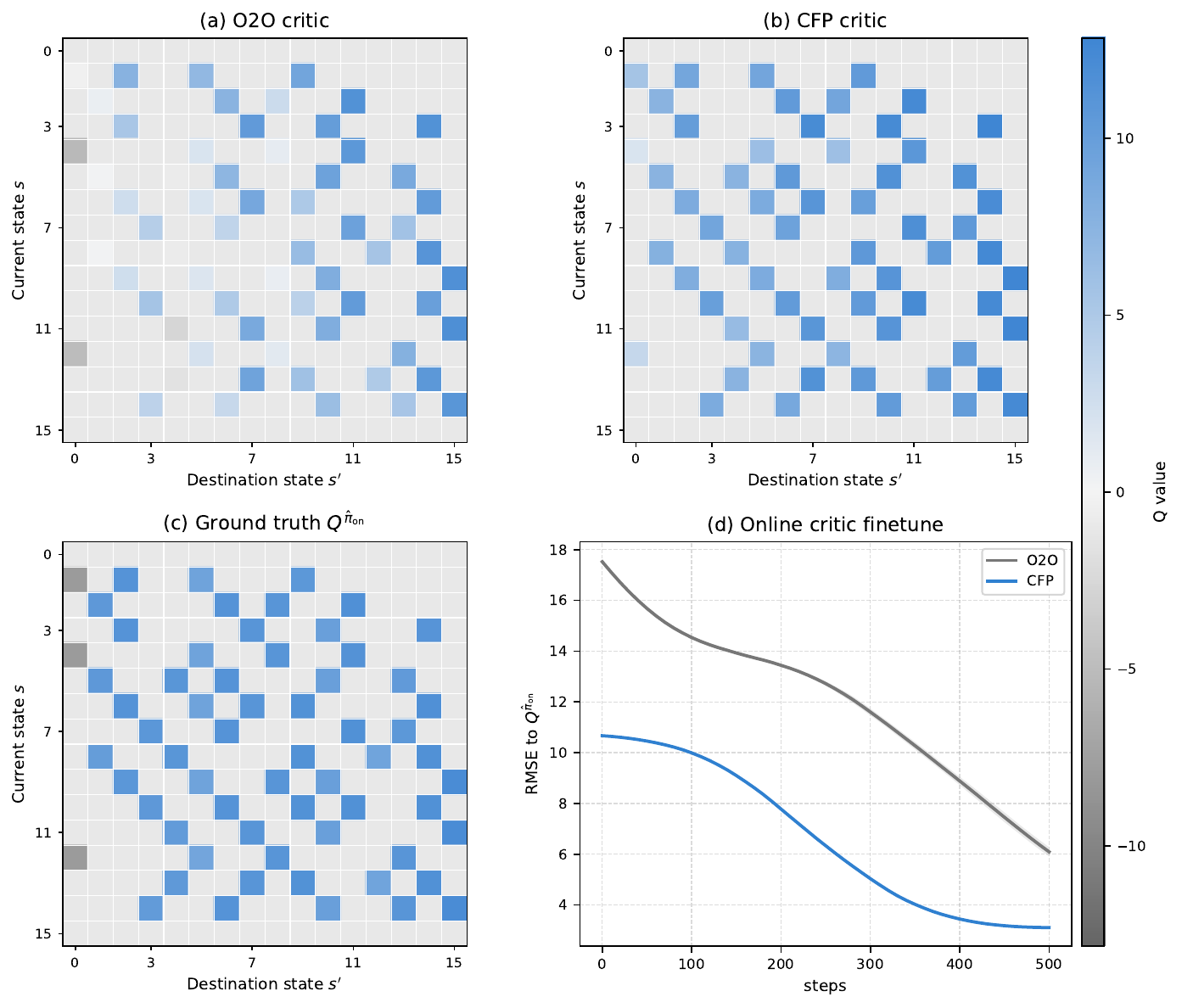}
    \caption{
        Critic calibration in the toy MDP. CFP produces Q value estimates closer to the ground truth $Q^{\pi_{\mathrm{on}}}$ and achieves substantially lower RMSE than O2O during online fine-tuning.
    }
    \label{fig:toy_exp}
\end{figure}

We evaluate each critic using the root mean squared error (RMSE) between the learned Q-values and the exact online Q-function $Q^{\hat\pi_{\mathrm{on}}}$ over all valid state transitions. As shown in Figure \ref{fig:toy_exp}, the CFP critic exhibits a Q-value structure that more closely resembles $Q^{\hat\pi_{\mathrm{on}}}$ than the O2O critic. Moreover, CFP maintains a lower RMSE throughout online fine-tuning than O2O. These results suggest that reusing the offline critic retains considerable bias, whereas initializing critic enables more accurate estimation of the online-policy value function. For more details, see Appendix \ref{App:toy_exp}.
\section{Related Work}
\label{headings}
\subsection{Offline-to-Online RL}
The primary objective of the O2O RL is to leverage offline pre-training on static datasets to enable more efficient fine-tuning during the subsequent online stage. This approach mitigates the fundamental challenges where training online RL from-scratch often fails to acquire effective policies in complex environments, thereby significantly reducing both computational overhead and time costs \citep{levine2020offline,xie2021policy, guo2023sample,pmlr-v202-yu23k,zhangpolicy,zhang2026sac}. However, O2O RL faces the problem of the distribution shift between the dataset and the environment, which causes catastrophic performance drop in online stage. To confront this problem, a few solutions are introduced. In this section, we list some of them that are related to our work.
\paragraph{Pessimism and Conservatism}
Early algorithms mainly focus on eliminating exploration errors. By augmenting the Bellman objective with a conservative regularizer, CQL
\citep{kumar2020conservative} lowers the estimated values of actions that are poorly supported by the offline dataset, thereby mitigating overestimation caused by distribution shift. Cal-QL \citep{nakamoto2023cal} constrains the learned conservative values to
remain above the value of a reference policy, thereby placing the Q-values on a reasonable scale for subsequent online fine-tuning. Some algorithms reach the same goal by explicitly restrict the learned policy from deviating too far from dataset distribution. For instance, BCQ \citep{fujimoto2019off} uses a conditional generative model and a perturbation model to restrict action selection to actions that are likely under, or close to, the dataset distribution. TD3+BC \citep{fujimoto2021minimalist} explicitly augments the actor objective with a behavior-cloning term, whereas AWAC \citep{nair2020awac} performs an advantage-weighted maximum-likelihood policy update, implicitly constraining the learned policy toward actions supported by the replay distribution.
\paragraph{Replay Design and Critic Calibration}
Beyond conservatism, recent offline-to-online RL methods improve fine-tuning through replay design and explicit value calibration. For example, RLPD \citep{ball2023efficient} trains an off-policy online RL agent with minibatches drawn jointly from fixed offline data and newly collected online experience, together with high update-to-data ratios and critic regularization for stable and sample-efficient learning. WSRL \citep{zhou2025efficient} introduces the warm-up stage in which only online data are used to calibrate the pretrained critic network. OPT \citep{shin2025online} introduces a newly initialized critic and an
additional online pre-training phase, during which the new critic is trained with both the offline dataset and a small set of online transitions. During fine-tuning, policy improvement is guided by a scheduled weighted combination of the offline-pretrained critic and the online-pretrained critic.

\section{Experiments}
\label{experiments}

To verify the effectiveness of CFP, we conduct extensive experiments on manipulation tasks under different environments. We further conducted ablation experiments, providing solid evidence for the implementation of the warm-up stage.
\subsection{Settings}
We conduct our experiments on 8 sparse reward domains, including 5 domains from OGBench \citep{ogbench_park2025}: \texttt{Cube-Double/Triple/Quadruple}, \texttt{Puzzle-4$\times$4}, \texttt{Scene}, and 3 tasks in Robomimic \citep{robomimic2021}: \texttt{Lift}, \texttt{Can} and \texttt{Square}. We use default \textbf{play} dataset for each OGBench domain. In particular, for \texttt{Cube Quadruple}, we use 100M-size dataset released by the OGBench authors \citep{ogbench_park2025}. In addition, we use the sparse-reward formulation for \texttt{Scene}. We use default \textbf{Multi-Human (MH)} dataset. For more details about the benchmark and dataset, see Appendix \ref{environment}.
\subsection{Baselines}
We select 4 representative flow-based algorithms as our baselines, including \textbf{(1) FQL} \citep{fql_park2025}, \textbf{(2) QC} \citep{ghasemipour2021emaq, li2026reinforcement}, \textbf{(3) QCFQL}  \citep{li2026reinforcement}, \textbf{(4) QCFQL-nstep}  \citep{li2026reinforcement,zhang2026sac}. We further implement CFP paradigm on all of them and compare the online performance among them. For more details, see Appendix \ref{App:baselines}.
\subsection{Main Result}
\begin{figure}[htbp]
    \centering
    \includegraphics[
        width=\textwidth
    ]{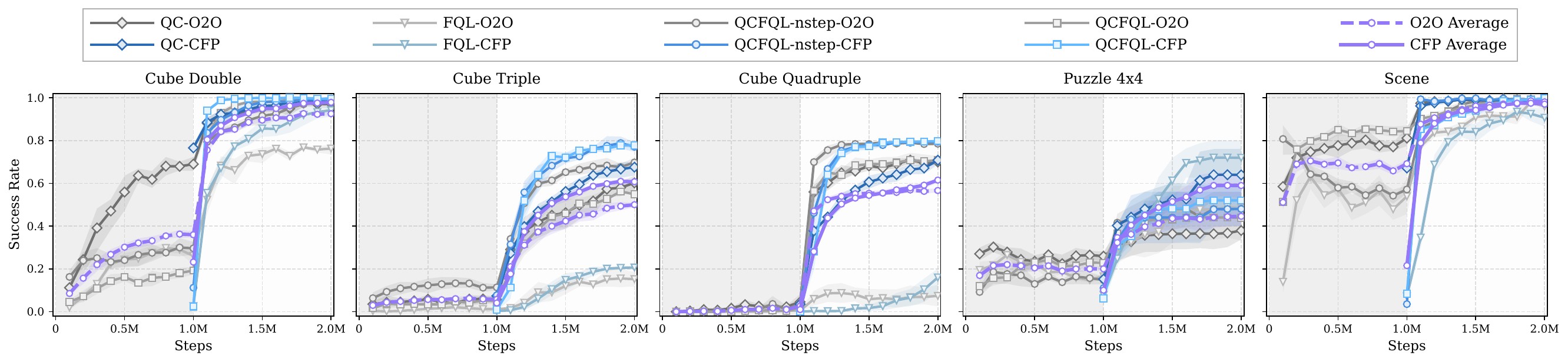}
    \caption{
        Performance across five tasks in each OGBench environment.
        Curves and shaded regions represent the mean and 95\%{} confidence
        interval across five random seeds, respectively.
    }
    \label{fig:ogbench-main-results}
\end{figure}

Figure \ref{fig:ogbench-main-results} reports the offline-to-online performance across five OGBench manipulation domains, with each curve aggregating results over five tasks. Overall, incorporating CFP consistently improves or preserves the performance of the underlying offline RL algorithms. The advantage of CFP is especially pronounced on Cube Triple, where all the CFP variants appear to outperform their corresponding O2O baselines. In particular, QCFQL-CFP and QCFQL-nstep-CFP exhibit remarkably similar learning dynamics across the five domains. Both variants rapidly improve after online fine-tuning begins and achieve nearly identical final performance on Cube Double, Cube Triple, Cube Quadruple, and Scene.

\begin{figure}[htbp]
    \centering
    \includegraphics[
        width=0.9\textwidth
    ]{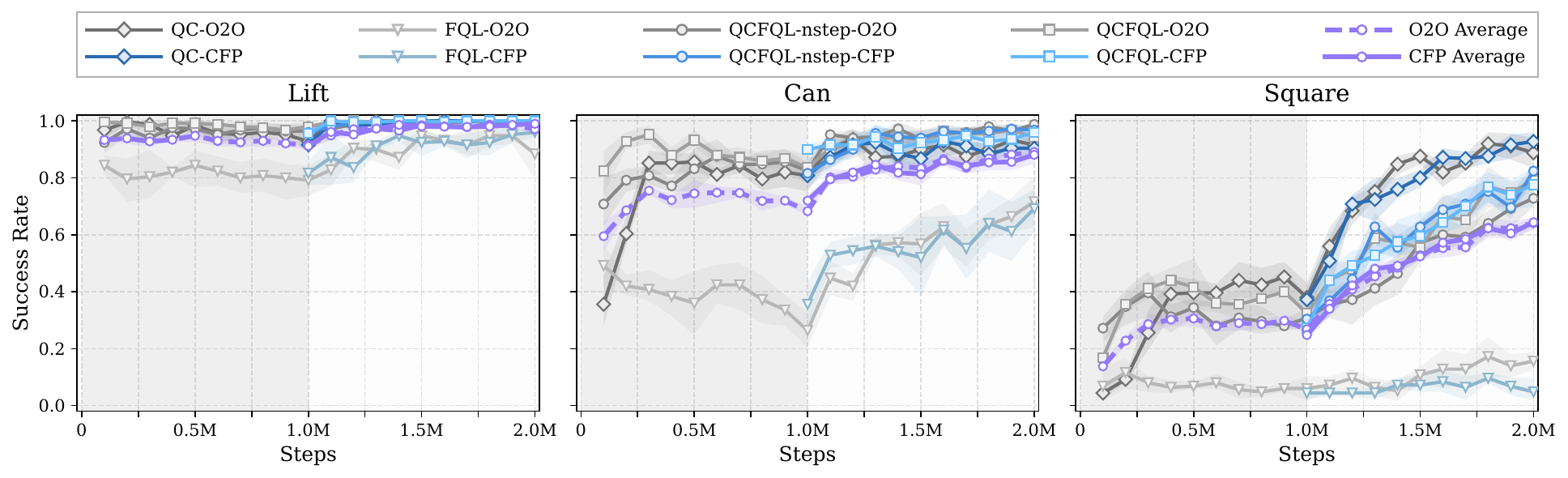}
    \caption{
        Performance on each Robomimic environment.
        Curves and shaded regions represent the mean and 95\%{} confidence
        interval across five random seeds, respectively.
    }
    \label{fig:robomimic-main-results}
\end{figure}
We also observe that the benefit of CFP depends on the underlying base algorithm. FQL-CFP achieves the largest improvement on Puzzle 4x4, reaching a final success rate of approximately $0.7$, compared with less than $0.4$ for FQL-O2O. Notably, FQL-CFP outperforms all QCFQL- and QC-based variants on this domain, despite being less competitive on Cube Triple and Cube Quadruple. This indicates that CFP can complement the inductive bias of FQL particularly well in environments requiring combinatorial manipulation, rather than producing a uniform improvement across all domains. QC-CFP, by contrast, yields more moderate but consistent gains, with its clearest improvements appearing on Cube Triple and Puzzle 4x4, while remaining comparable to QC-O2O on Cube Double, Cube Quadruple, and Scene. Taken together, these results suggest that CFP is broadly compatible with different offline RL backbones, while the magnitude of its benefit is jointly determined by the base algorithm and the structural characteristics of the downstream environment.

Figure \ref{fig:robomimic-main-results} reports the offline-to-online performance on three Robomimic manipulation domains. Overall, the CFP variants achieve performance comparable to their corresponding O2O baselines across most algorithms and environments. The main exception is QCFQL-nstep-CFP on Square, which exhibits a substantial improvement over QCFQL-nstep-O2O and achieves a higher final success rate. For full results, see Appendix \ref{App:full_result}.
\subsection{Ablation study}
In our ablation studies, we aim to answer the following questions:\begin{itemize}\item \textbf{How should the warm-up stage be designed?}\item \textbf{How sensitive is CFP to the number of warm-up steps?}\end{itemize}
\subsubsection{Should we allow the fresh critic to guide actor?}
Since the fresh critic in CFP is initially untrained, a natural concern is that the gradients propagated through the Q-loss, $L_{Q}=\mathbb{E}_{s\sim \mathcal{D},a\sim \pi_{\omega}(\cdot|s)}[-Q_\phi(s,a)] $, may be uninformative or even harmful to the pretrained actor. This consideration suggests that, during the warm-up stage, the actor should continue to be optimized solely with the behavior cloning objective. Our experiments, however, lead to the opposite conclusion. We find that, despite the short duration of the warm-up stage---only $0.5\%  $ of the total training steps---the critic converges rapidly and already exerts a substantial influence on the actor. To investigate this effect, we compare online fine-tuning performance with and without incorporating the Q-function loss into the actor update during warm-up\footnote{Among the algorithms considered in this paper, only QC does not apply backpropagation; therefore, the discussion in this section is restricted to the other three algorithms.}.
\begin{figure}[H]
    \centering
    \includegraphics[
        width=\textwidth
    ]{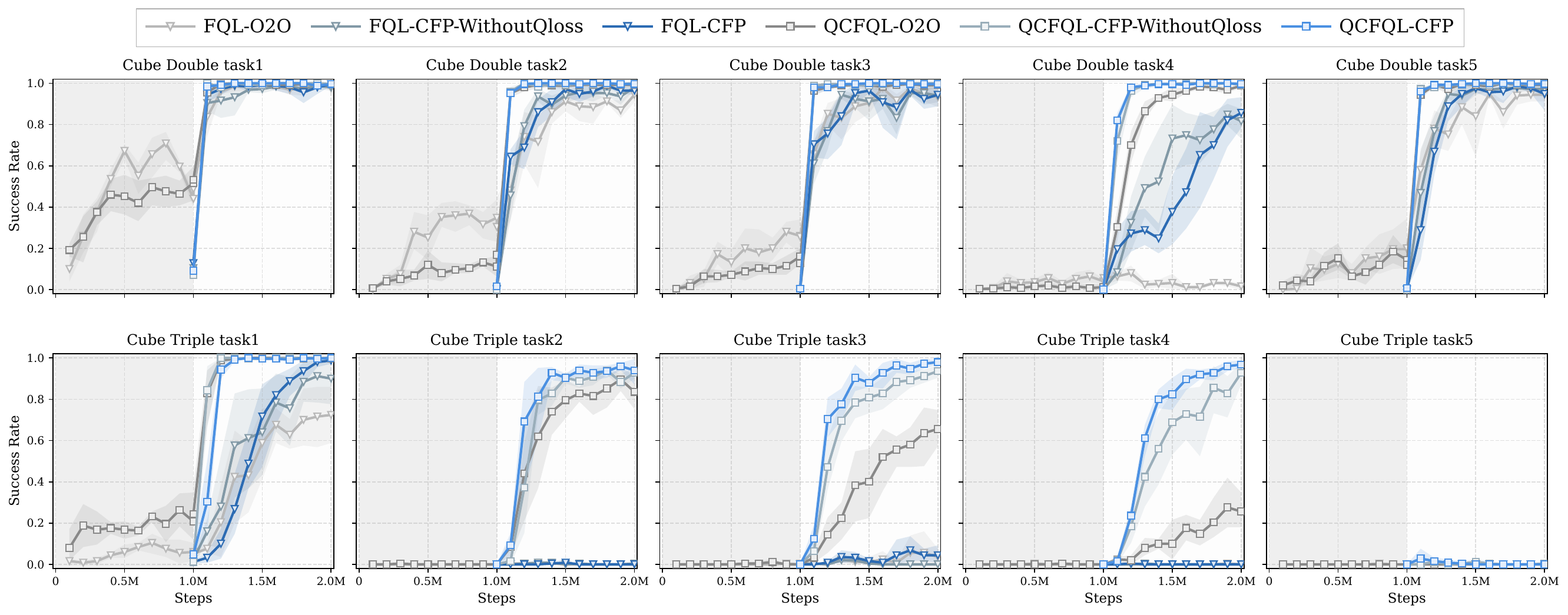}
    \caption{
        Ablation of the Q-loss during warm-up on OGBench. Curves and shaded regions represent the mean and 95\%{} confidence
        interval across five random seeds, respectively.
    }
    \label{fig:ablation_qloss}
\end{figure}

Figure \ref{fig:ablation_qloss} suggests that the Q-loss backpropagating through flow networks during warm-up generally improves the subsequent online fine-tuning performance, with the clearest gains appearing on the more challenging Cube Triple tasks and Cube Double task 4.

\subsubsection{Does the number of warm-up steps matter?}
\label{sec:warm-up steps}
We view the number of steps during warm-up stage as a hyperparameter. We sweep this hyperparameter from 0 to 0.1M to see whether CFP is sensitive to it.

\begin{figure}[H]
    \centering
    \includegraphics[
        width=\textwidth
    ]{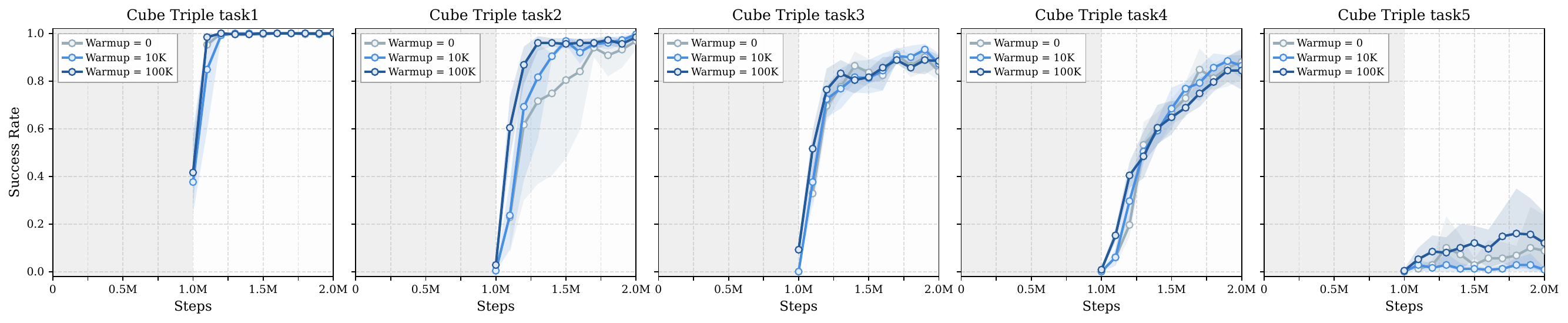}
    \caption{
        Ablation of the number of steps during warm-up on OGBench. The applied algorithm is QCFQL-nstep-CFP. Curves and shaded regions represent the mean and 95\%{} confidence
        interval across five random seeds, respectively.
    }
    \label{fig:ablation_steps}
\end{figure}

Figure \ref{fig:ablation_steps} suggests that CFP is relatively insensitive to the length of warm-up. Warm-up-free CFP may experience unstable training during online fine-tuning. Hence, to balance stability and computation efficiency, we choose 10K as our main hyperparameter. We process our training on NVIDIA A800 GPU. In the absence of other concurrent workloads, a 10K-step warm-up requires only approximately 30 seconds, making CFP computationally efficient. For more results of ablation study, see Appendix \ref{App:ablation study}.
\section{Conclusion and Discussion}
\label{Discussionsandconclusions}
We re-examined the training process of the Actor-Critic framework within the O2O paradigm. Our research reveals that the asymmetric training of the two components in this framework is key to achieving efficient online fine-tuning. Specifically, while the offline phase provides the actor with a sampling prior---thereby enhancing sampling efficiency during the online phase---it simultaneously introduces potential pessimism and distribution-shift-induced mismatches for the critic. We compared the mean critic outputs between O2O and CFP and designed an MDP to validate our inferences. These findings offer new insights into the role of datasets in RL training process.

However, CFP does not consistently outperform conventional O2O training on Robomimic. This limitation suggests that discarding the offline critic is most beneficial when inherited value estimates exhibit substantial distribution-induced mismatch. Developing diagnostics that predict when such mismatch occurs, and designing more effective critic-initialization and warm-up strategies, are promising directions for future work.

\bibliography{iclr2025_conference}
\bibliographystyle{iclr2025_conference}

\appendix

\section{Environment}
\label{environment}
In this section, we present the domain we use in our experiments in detail. Table \ref{tab:environment_specifications} summarizes some information of each domain, including \textbf{Reward Type, Dataset Size, Episode Length and Action Dimension.}
\begin{figure}[htbp]
    \centering

    \begin{subfigure}[b]{0.24\textwidth}
        \centering
        \includegraphics[width=\linewidth]{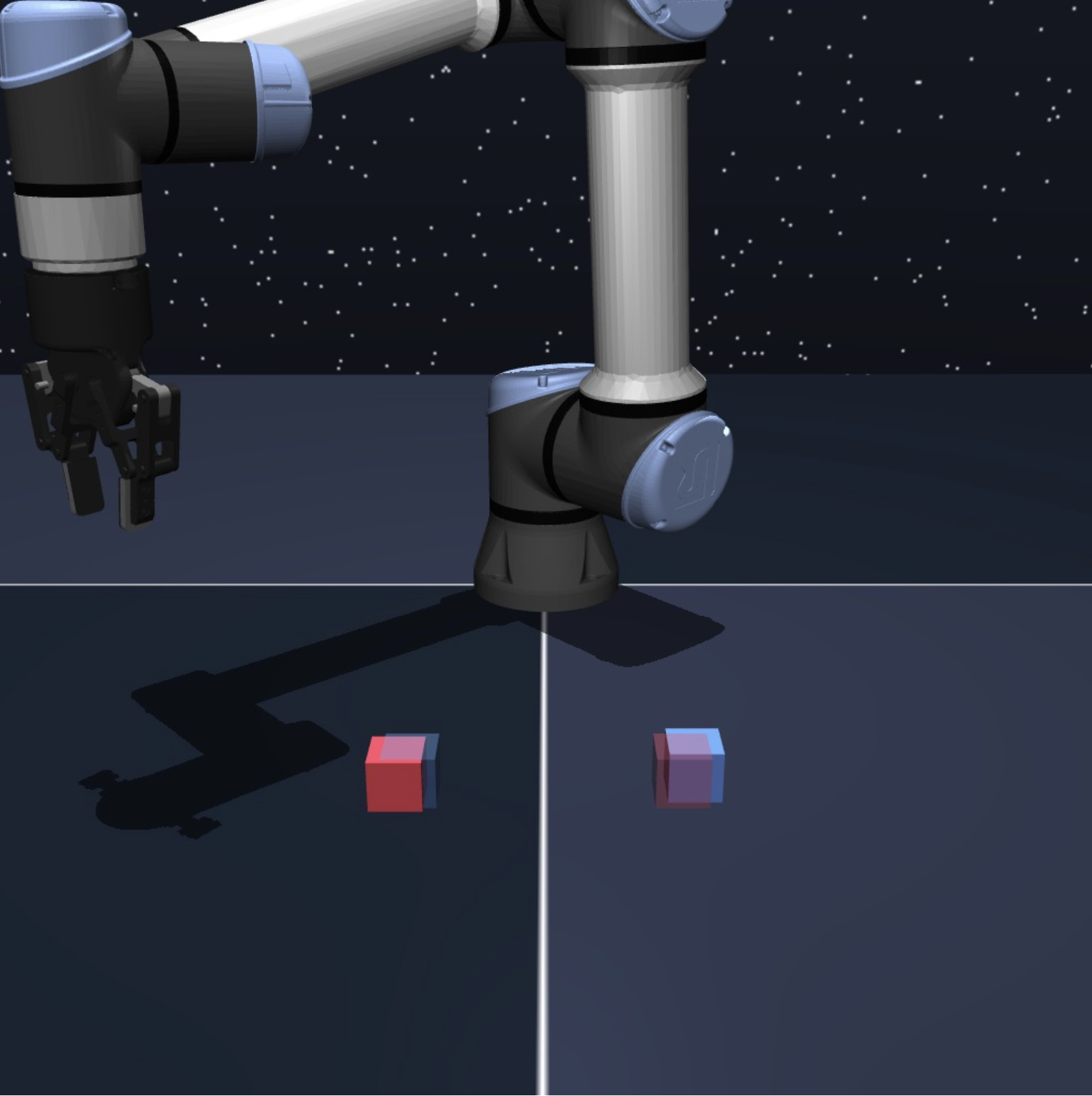}
        \caption{Cube Double}
        \label{fig:env1}
    \end{subfigure}\hfill % \hfill 用于自动填充水平间距
    \begin{subfigure}[b]{0.24\textwidth}
        \centering
        \includegraphics[width=\linewidth]{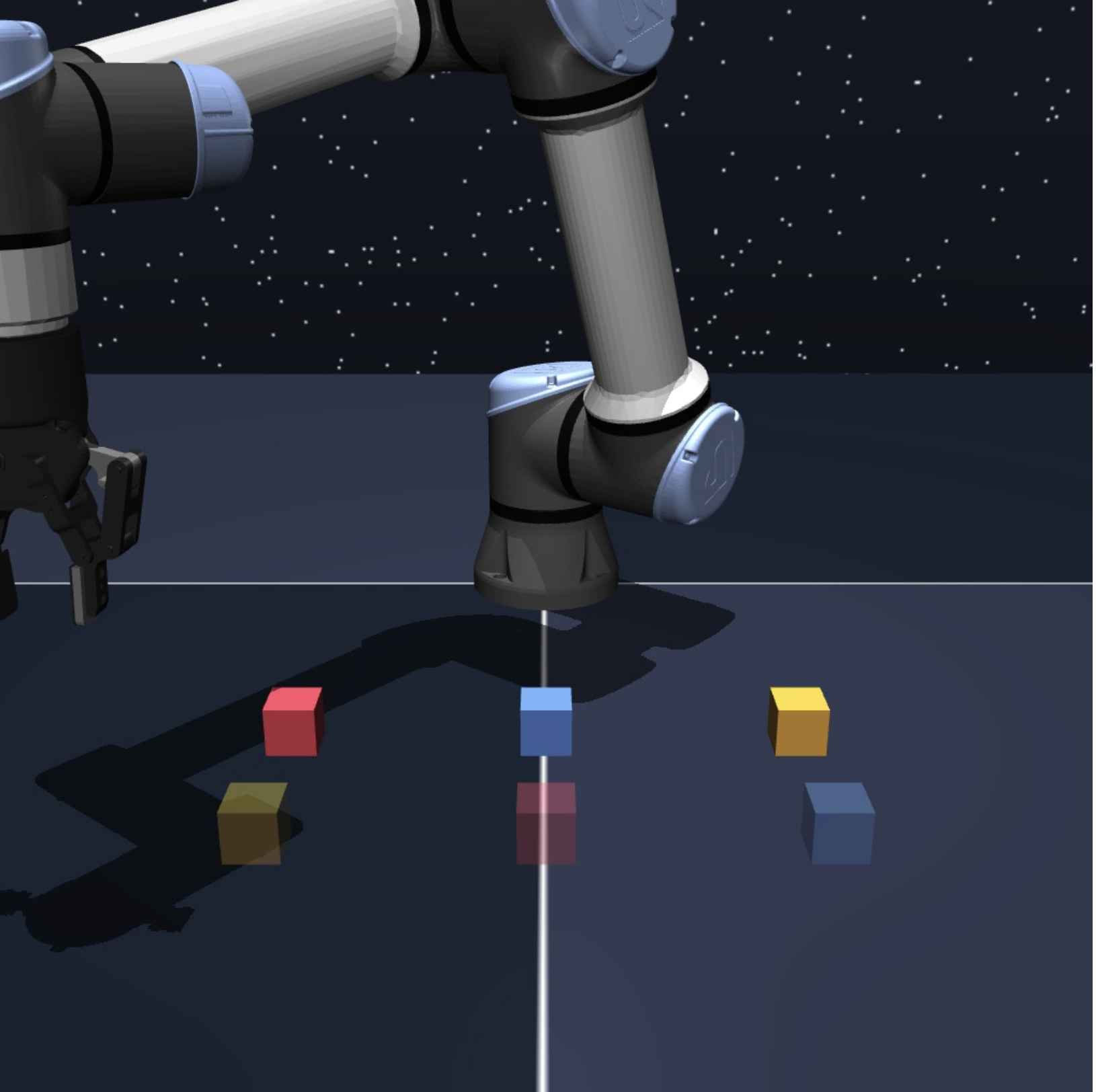}
        \caption{Cube Triple}
        \label{fig:env2}
    \end{subfigure}\hfill
    \begin{subfigure}[b]{0.24\textwidth}
        \centering
        \includegraphics[width=\linewidth]{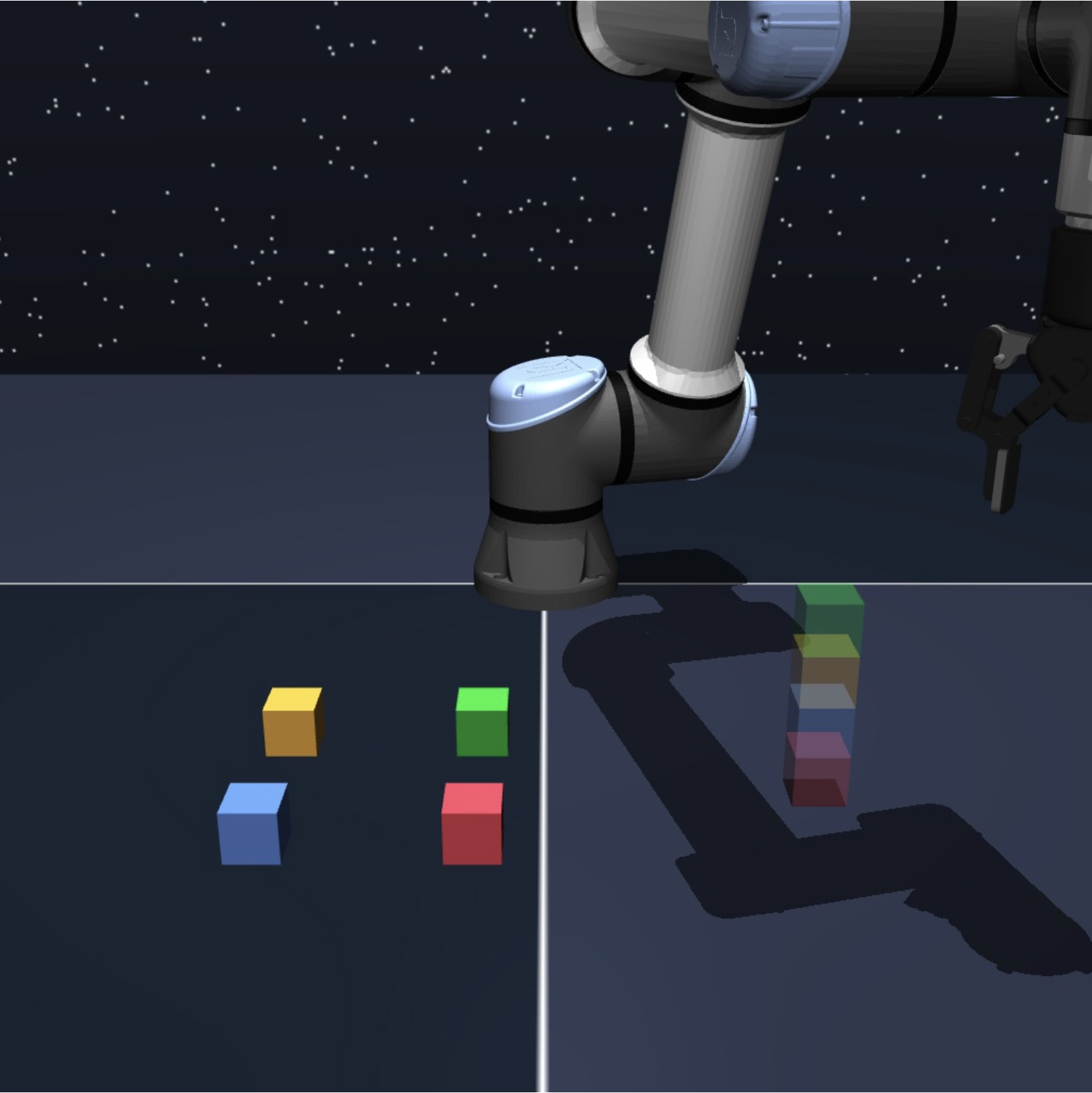}
        \caption{Cube Quadruple}
        \label{fig:env3}
    \end{subfigure}\hfill
    \begin{subfigure}[b]{0.24\textwidth}
        \centering
        \includegraphics[width=\linewidth]{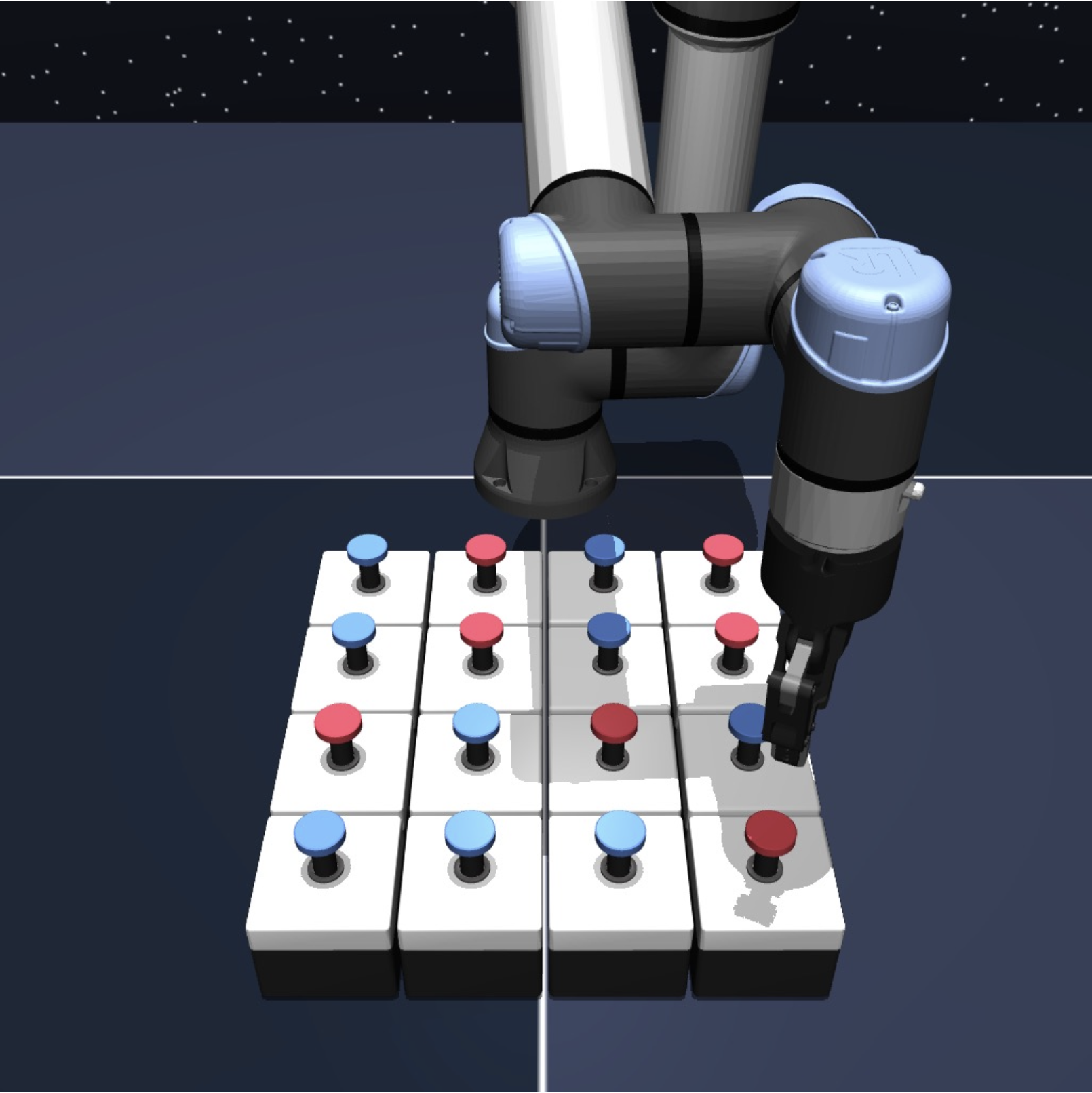}
        \caption{Puzzle $4\times4$}
        \label{fig:env4}
    \end{subfigure}
    
    \vspace{1em} 
    
    \begin{subfigure}[b]{0.24\textwidth}
        \centering
        \includegraphics[width=\linewidth]{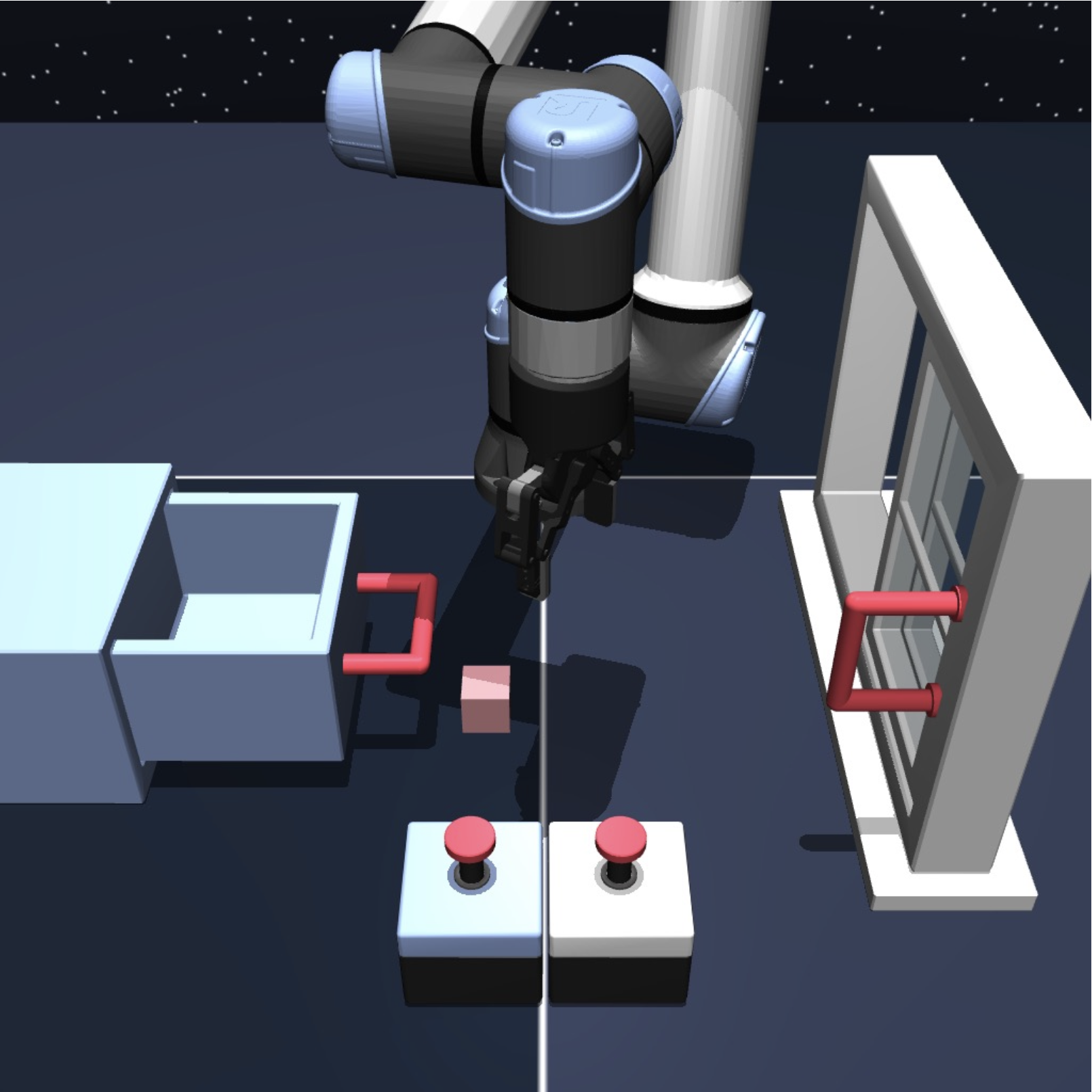}
        \caption{Scene}
        \label{fig:env5}
    \end{subfigure}\hfill
    \begin{subfigure}[b]{0.24\textwidth}
        \centering
        \includegraphics[width=\linewidth]{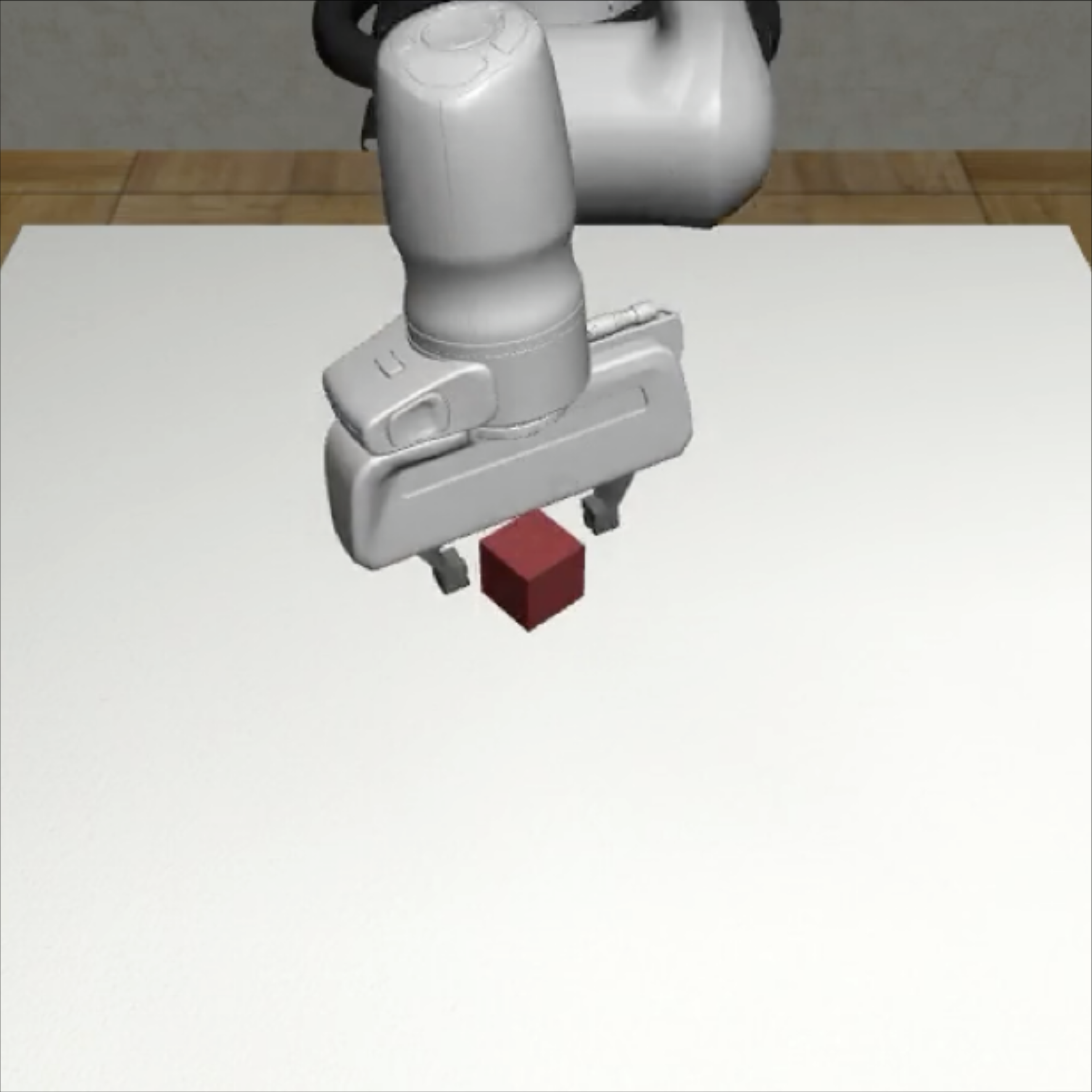}
        \caption{Lift}
        \label{fig:env6}
    \end{subfigure}\hfill
    \begin{subfigure}[b]{0.24\textwidth}
        \centering
        \includegraphics[width=\linewidth]{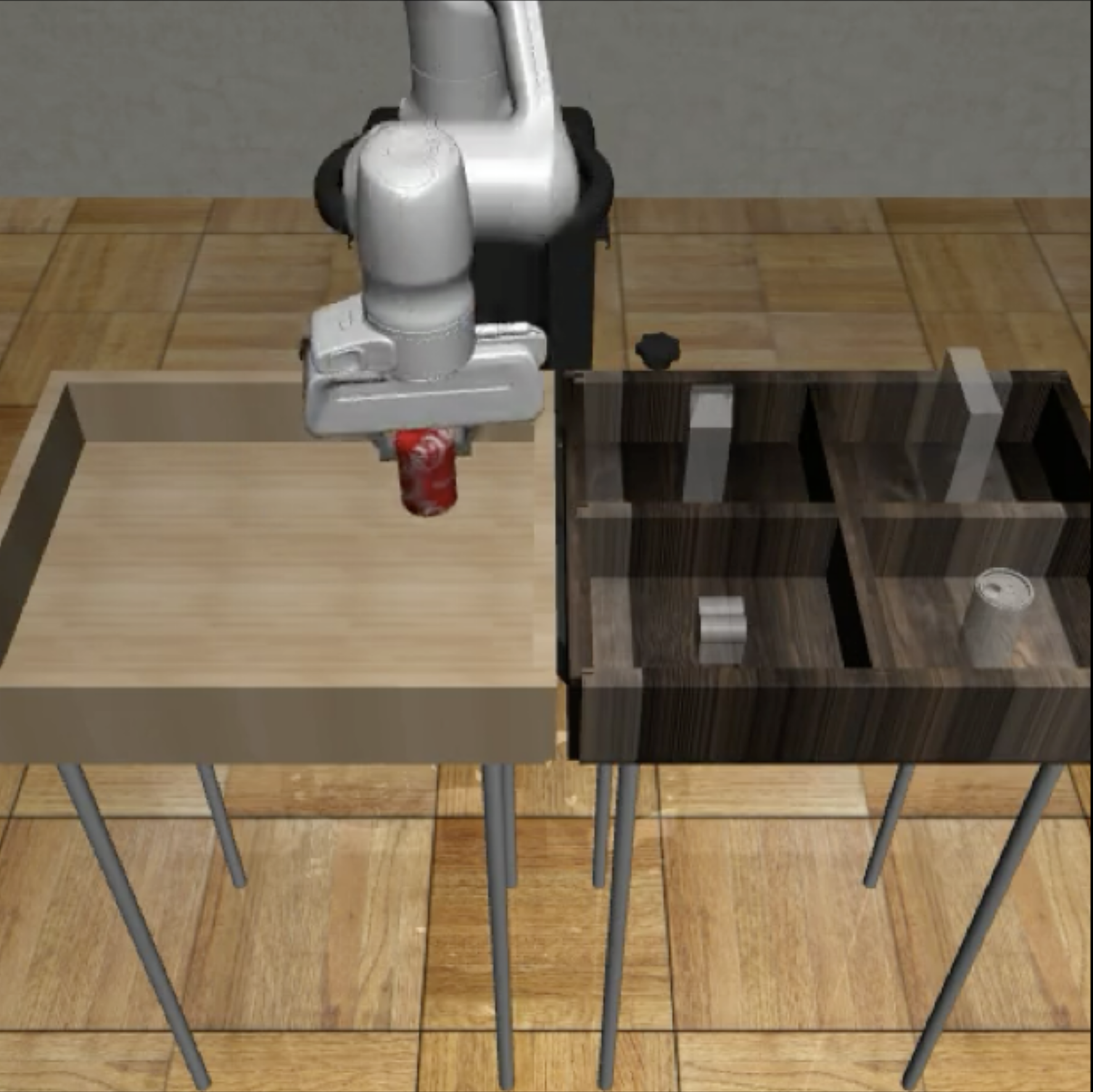}
        \caption{Can}
        \label{fig:env7}
    \end{subfigure}\hfill
    \begin{subfigure}[b]{0.24\textwidth}
        \centering
        \includegraphics[width=\linewidth]{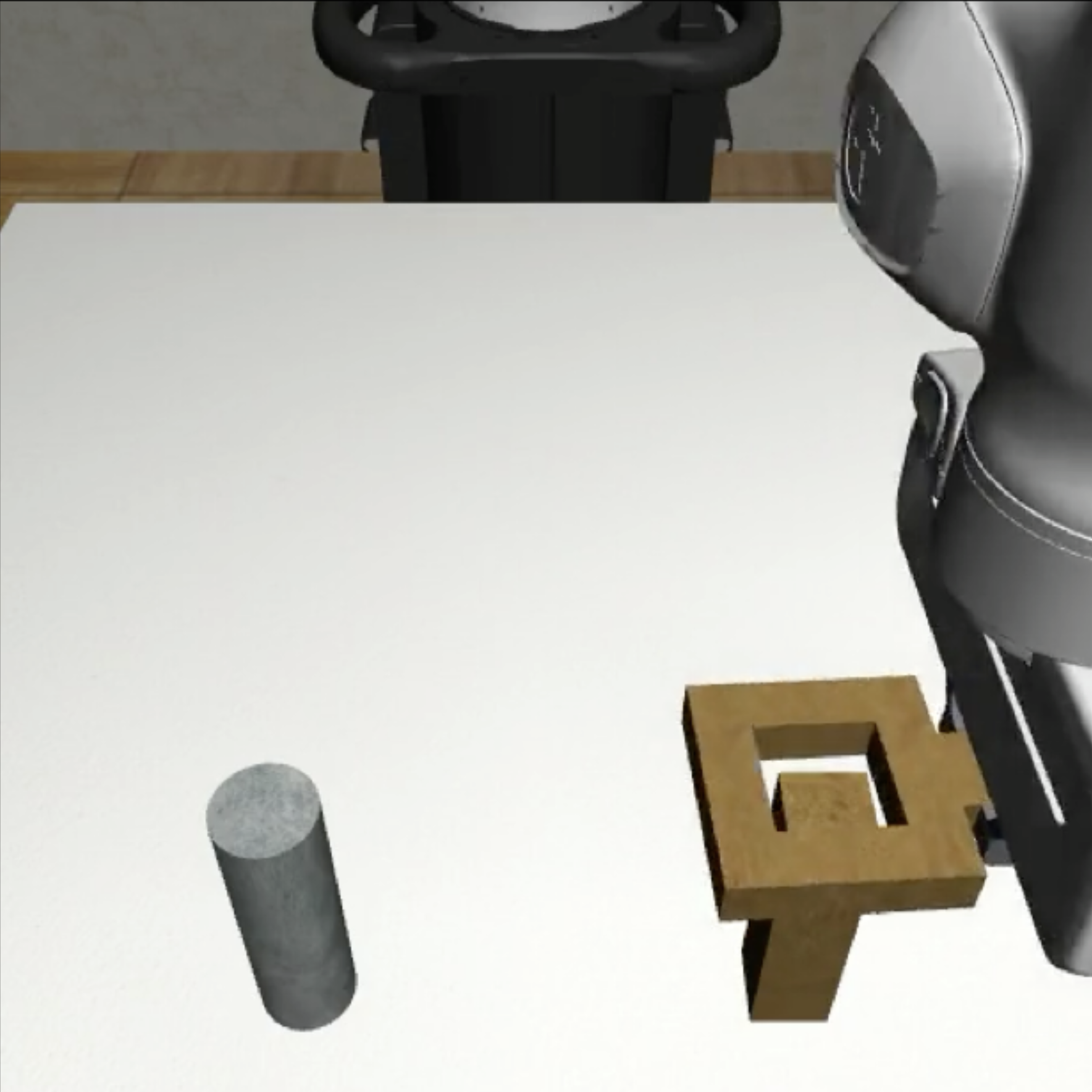}
        \caption{Square}
        \label{fig:env8}
    \end{subfigure}

    \caption{
        Visualizations of the environments. Subfigures (a-e) show the OGBench domains. Subfigures (f-h) depict the manipulation tasks from Robomimic.
    }
    \label{fig:environments}
\end{figure}
\subsection{OGBench}
We choose 5 OGBench manipulation domains and use the single-task version of it. We use the standard \textbf{play} version of the dataset, in which trajectories were collected using a non-Markovian expert policy. For \texttt{Cube Double/Triple/Quadruple}, the agents need to manipulate an arm and learn a pick-and-place skill to place colored cube blocks to the right locations. Particularly, \texttt{Cube Quadruple} is an extremely challenging domain in which the agent must process four cubes in sequence. It is almost impossible to learn an effective policy under standard dataset. Hence, we use the 100M-size dataset provided officially. \texttt{Puzzle 4$\times$4} requires solving the "Lights Out" puzzle with a robot arm. For \texttt{Scene-sparse}, we use the sparse-reward mode of the reward function such that the agent receives a reward of 0 only upon completing the task, and -1 in all other cases.

\begin{table}[htbp]
    \centering
    \caption{Specifications of the environments used in our experiments.}
    \label{tab:environment_specifications}
    \resizebox{\textwidth}{!}{%
    \begin{tabular}{@{}lcccc@{}}
        \toprule
        \textbf{Environment}
        & \textbf{Reward Type}
        & \textbf{Dataset Size}
        & \textbf{Episode Length}
        & \textbf{Action Dimension} \\
        \midrule

        \multicolumn{5}{@{}l}{\textit{OGBench}} \\
        \midrule
        cube-double    & Sparse & 1M & 500  & 5 \\
        cube-triple    & Sparse & 3M & 1000 & 5 \\
        cube-quadruple & Sparse & 100M & 1000 & 5 \\
        puzzle-4x4     & Sparse & 1M & 500   & 5 \\
        scene          & Sparse & 1M & 750   & 5 \\

        \midrule
        \multicolumn{5}{@{}l}{\textit{Robomimic}} \\
        \midrule
        lift   & Sparse & 31,127 & 500 & 7 \\
        can    & Sparse & 62,756 & 500 & 7 \\
        square & Sparse & 80,731 & 500 & 7 \\

        \bottomrule
    \end{tabular}%
    }
\end{table}
\subsection{Robomimic}
We select 3 challenging domain from Robomimic, including \texttt{Lift, Can and Square}. We conduct our experiments on Multi-Human datasets. These datasets were collected by 6 operators, including 300 successful trajectories. \texttt{Lift} requires the robot arm to pick up a cube. \texttt{Can} requires the robot arm to pick up a can and place it into a container. \texttt{Square} requires the robot arm to pick up a square sleeve and fit it onto a square post.
\section{Implementation}
\label{App:baselines}
In this section, we present the details of our implementation of CFP on all the baselines.

\paragraph{Flow Q-Learning.}
FQL is a strong baseline for offline reinforcement learning. Its main innovation is to exploit the expressive power of a multi-step flow network to approximate the behavior distribution in the offline dataset, while using a Q-loss to propagate gradients from the critic to the actor. Directly backpropagating through the multi-step flow generation process, however, may lead to unstable gradients. To avoid this issue, FQL introduces a one-step noise-conditioned policy that directly distills the actions generated by the multi-step behavior flow. Although the one-step policy may sacrifice some of the expressive capacity of the original multi-step flow, it enables more stable policy optimization.

More concretely, FQL consists of the following three networks:
\begin{enumerate}
    \item a behavior flow policy \(\mu_{\theta}(s,z)\), which generates an action from a state \(s\) and Gaussian noise \(z\sim\mathcal{N}(0,I^d)\) through multi-step flow integration;
    \item a one-step noise-conditioned policy \(\mu_{\omega}(s,z)\), which distills the output of the behavior flow policy; and
    \item a critic \(Q_{\phi}(s,a)\), which estimates the expected return of a state--action pair.
\end{enumerate}

The one-step policy is optimized by combining flow-policy distillation with Q-value maximization:
\begin{equation}
L_{\pi}(\omega)
=
\alpha
\mathbb{E}_{
    \substack{
        s\sim\mathcal{D},\\
        z\sim\mathcal{N}(0,I^d)
    }
}
\left[
    \left\|
        \mu_{\omega}(s,z)
        -   
        \mu_{\theta}(s,z)
    \right\|_2^2
\right]+
\mathbb{E}_{
    \substack{
        s\sim\mathcal{D},\\
        a\sim\pi_{\omega}(\cdot\mid s)
    }
}
\left[
    -Q_{\phi}(s,a)
\right],
\label{eq:fql_actor_loss}
\end{equation}
where \(\alpha\) controls the strength of behavior regularization. The first term distills the output of the multi-step behavior flow into the one-step policy, whereas the second term propagates the critic gradient to the one-step actor.

The critic is trained with the temporal-difference objective
\begin{equation}
L_{Q}(\phi)
=
\mathbb{E}_{(s,a,r,s')\sim\mathcal{D}}
\left[
    \left(
        Q_{\phi}(s,a)
        -
        \left[
            r
            +
            \gamma
            Q_{\bar{\phi}}(s',a')
        \right]
    \right)^2
\right],
\label{eq:fql_critic_loss}
\end{equation}
where \(a'\sim\pi_{\omega}(\cdot\mid s')\) and \(Q_{\bar{\phi}}\) denotes the target critic.

\paragraph{QCFQL.}
QCFQL extends FQL to temporally extended action spaces by jointly generating and evaluating an action chunk
\(
\mathbf{a}_t=(a_t,\ldots,a_{t+h-1})
\). The QCFQL actor objective is
\begin{equation}
L_{\pi}(\omega)
=
\alpha
\mathbb{E}_{
    \substack{
        s_t\sim\mathcal{D},\\
        z\sim\mathcal{N}(0,I^{hd})
    }
}
\left[
    \left\|
        \mu_{\omega}(s_t,z)
        -
        \mu_{\theta}(s_t,z)
    \right\|_2^2
\right]+
\mathbb{E}_{
    \substack{
        s_t\sim\mathcal{D},\\
        \mathbf{a}_t\sim\pi_{\omega}(\cdot\mid s_t)
    }
}
\left[
    -Q_{\phi}(s_t,\mathbf{a}_t)
\right].
\label{eq:qcfql_actor_loss}
\end{equation}

The QCFQL critic is trained using an \(h\)-step temporal-difference target:
\begin{equation}
L_{Q}(\phi)
=
\mathbb{E}
\left[
    \left(
        Q_{\phi}(s_t,\mathbf{a}_t)
        -
        \left[
            r_t^{(h)}
            +
            \gamma^h
            Q_{\bar{\phi}}
            (s_{t+h},\mathbf{a}_{t+h})
        \right]
    \right)^2
\right],
\label{eq:qcfql_critic_loss}
\end{equation}
where
\begin{equation}
r_t^{(h)}
=
\sum_{j=0}^{h-1}\gamma^j r_{t+j},
\qquad
\mathbf{a}_{t+h}
\sim
\pi_{\omega}(\cdot\mid s_{t+h}).
\end{equation}

\paragraph{QCFQL-nstep.}
QCFQL-nstep is a similar implementation of QCFQL. Instead of distilling the multi-step network with single-step network, we use direct backpropagation. The QCFQL-nstep actor objective is
\begin{equation}
L_{\pi}(\theta)
=
\alpha
\mathbb{E}_{
    \substack{
        (s_t,\mathbf{a}_t)\sim\mathcal{D},\\
        z\sim\mathcal{N}(0,I^{hd})
    }
}
\left[
    \left\|
        \mu_{\theta}(s_t,z) - \mathbf{a}_t
    \right\|_2^2
\right]+
\mathbb{E}_{
    \substack{
        s_t\sim\mathcal{D},\\
        \mathbf{a}_t\sim\pi_{\theta}(\cdot\mid s_t)
    }
}
\left[
    -Q_{\phi}(s_t,\mathbf{a}_t)
\right].
\label{eq:qcfql_nstep_actor_loss}
\end{equation}
The remaining components of QCFQL-nstep are identical to those of QCFQL.
\paragraph{QC.}
QC uses the same flow-matching behavior policy
\(\mu_{\theta}\) as FQL, but does not introduce an additional one-step noise-conditioned actor. Instead, QC implicitly defines its policy through best-of-\(N\) candidate selection. Specifically, for a given state \(s\), we first sample \(N\) independent Gaussian noise vectors,
\begin{equation}
    z^1,z^2,\ldots,z^N
    \overset{\mathrm{i.i.d.}}{\sim}
    \mathcal{N}(0,I^{hd}),
\end{equation}
and generate \(N\) candidate actions using the multi-step behavior flow:
\begin{equation}
    \mathbf{a}_t^i=\mu_{\theta}(s_t,z^i),
    \qquad i\in\{1,\ldots,N\}.
    \label{eq:qc_candidates}
\end{equation}
The policy then selects the candidate with the largest estimated Q-value:
\begin{equation}
    i^*
    =
    \arg\max_{i\in\{1,\ldots,N\}}
    Q_{\phi}(s_t,\mathbf{a}^i),
    \qquad
    \pi_{\mathrm{QC}}(s)=\mathbf{a}^{i^*}.
    \label{eq:qc_selection}
\end{equation}

Unlike FQL and QCFQL, QC does not backpropagate the critic gradient. The selected best-of-\(N\) action is directly used to construct the temporal-difference target. Accordingly, the critic is optimized by
\begin{equation}
L_{Q}(\phi)
=
\mathbb{E}_{(s,\mathbf{a}_t,r,s')\sim\mathcal{D}}
\left[
    \left(
        Q_{\phi}(s_t,\mathbf{a_t})
        -
        \left[
            r_t^{(h)}
            +
            \gamma^h
            Q_{\bar{\phi}}
            \left(
                s_{t+h},
                \mathbf{a}_{t+h}^{\,i^*}
            \right)
        \right]
    \right)^2
\right],
\label{eq:qc_critic_loss}
\end{equation}
where $\mathbf{a^{\,i^*}_{t+h}}\sim \pi_{QC}(\cdot|s_{t+h})$.

Next, we demonstrate how we incorporate CFP into the original algorithm.
\paragraph{FQL-CFP.}
Since the implementations of all the algorithms are largely similar, we primarily describe how FQL-CFP is implemented. 

The objective of FQL-CFP during the offline stage is:

\begin{equation}
L_{\pi}(\omega)
=
\alpha
\mathbb{E}_{
    \substack{
        s\sim\mathcal{D},\\
        z\sim\mathcal{N}(0,I^d)
    }
}
\left[
    \left\|
        \mu_{\omega}(s,z)
        -   
        \mu_{\theta}(s,z)
    \right\|_2^2
\right].
\end{equation}
CFP does not present critic updates during the offline stage.

During the warm-up phase, the actor and the initialized critic undergo brief updates and calibration using the offline dataset, with the following update objective: 
\begin{equation}
L_{\pi}(\omega)
=
\alpha
\mathbb{E}_{
    \substack{
        s\sim\mathcal{D},\\
        z\sim\mathcal{N}(0,I^d)
    }
}
\left[
    \left\|
        \mu_{\omega}(s,z)
        -   
        \mu_{\theta}(s,z)
    \right\|_2^2
\right]-\mathbb E_{s\sim\mathcal D,\,
a\sim\pi_\omega(\cdot|s)}
[Q_\phi(s,a)].
\end{equation}

The critic is trained with the temporal-difference objective
\begin{equation}
L_{Q}(\phi)
=
\mathbb{E}_{(s,a,r,s')\sim\mathcal{D}}
\left[
    \left(
        Q_{\phi}(s,a)
        -
        \left[
            r
            +
            \gamma
            Q_{\bar{\phi}}(s',a')
        \right]
    \right)^2
\right],
\label{eq:fqlcfp_critic_loss}
\end{equation}
where \(a'\sim\pi_\omega(\cdot\mid s')\) and terminal transitions are
masked in the bootstrap target. During the online phase, the agent samples trajectories in the real environment and stores them in the batch. The objectives are:

\begin{equation}
L_{\pi}(\omega)
=
\alpha
\mathbb{E}_{
    \substack{
        s\sim\mathcal{B},\\
        z\sim\mathcal{N}(0,I^d)
    }
}
\left[
    \left\|
        \mu_{\omega}(s,z)
        -   
        \mu_{\theta}(s,z)
    \right\|_2^2
\right]-\mathbb E_{s\sim\mathcal D,\,
a\sim\pi_\omega(\cdot|s)}
[Q_\phi(s,a)],
\end{equation}

\begin{equation}
L_{Q}(\phi)
=
\mathbb{E}_{(s,a,r,s')\sim\mathcal{B}}
\left[
    \left(
        Q_{\phi}(s,a)
        -
        \left[
            r
            +
            \gamma
            Q_{\bar{\phi}}(s',a')
        \right]
    \right)^2
\right].
\end{equation}

Since the objectives for the warm-up and online phases are identical—differing only in the source of the updates (the warm-up phase relies entirely on the dataset, whereas the online phase utilizes batches containing online data)---no further distinction will be made between the two.

\paragraph{QCFQL-CFP.}
The update method is very similar to that of FQL-CFP; QCFQL simply incorporates an action chunk. Its update objective during the offline stage is
\begin{equation}
L_{\pi}(\omega)
=
\alpha
\mathbb{E}_{
    \substack{
        s_t\sim\mathcal{D},\\
        z\sim\mathcal{N}(0,I^{hd})
    }
}
\left[
    \left\|
        \mu_{\omega}(s_t,z)
        -
        \mu_{\theta}(s_t,z)
    \right\|_2^2
\right]
\label{eq:qcfqlcfp_actor_loss}
\end{equation}
For both warm-up and online stage, the targets are:
\begin{equation}
L_{\pi}(\omega)
=
\alpha
\mathbb{E}_{
    \substack{
        s_t\sim\mathcal{D},\\
        z\sim\mathcal{N}(0,I^{hd})
    }
}
\left[
    \left\|
        \mu_{\omega}(s_t,z)
        -
        \mu_{\theta}(s_t,z)
    \right\|_2^2
\right]
+
\mathbb{E}_{
    \substack{
        s_t\sim\mathcal{D},\\
        \mathbf{a}_t\sim\pi_{\omega}(\cdot\mid s_t)
    }
}
\left[
    -Q_{\phi}(s_t,\mathbf{a}_t)
\right],
\end{equation}
\begin{equation}
L_{Q}(\phi)
=
\mathbb{E}
\left[
    \left(
        Q_{\phi}(s_t,\mathbf{a}_t)
        -
        \left[
            r_t^{(h)}
            +
            \gamma^h
            Q_{\bar{\phi}}
            (s_{t+h},\mathbf{a}_{t+h})
        \right]
    \right)^2
\right].
\end{equation}

\paragraph{QCFQL-nstep-CFP.}
QCFQL-nstep-CFP has a similar implementation. The QCFQL-nstep actor objective during offline phase is:
\begin{equation}
L_{\pi}(\theta)
=
\alpha
\mathbb{E}_{
    \substack{
        (s_t,\mathbf{a}_t)\sim\mathcal{D},\\
        z\sim\mathcal{N}(0,I^{hd})
    }
}
\left[
    \left\|
        \mu_{\theta}(s_t,z) - \mathbf{a}_t
    \right\|_2^2
\right]
\end{equation}
For warm-up and online updates, the objectives are:
\begin{equation}
L_{\pi}(\theta)
=
\alpha
\mathbb{E}_{
    \substack{
        (s_t,\mathbf{a}_t)\sim\mathcal{D},\\
        z\sim\mathcal{N}(0,I^{hd})
    }
}
\left[
    \left\|
        \mu_{\theta}(s_t,z) - \mathbf{a}_t
    \right\|_2^2
\right]
+
\mathbb{E}_{
    \substack{
        s_t\sim\mathcal{D},\\
        \mathbf{a}_t\sim\pi_{\theta}(\cdot\mid s_t)
    }
}
\left[
    -Q_{\phi}(s_t,\mathbf{a}_t)
\right],
\end{equation}
\begin{equation}
L_{Q}(\phi)
=
\mathbb{E}
\left[
    \left(
        Q_{\phi}(s_t,\mathbf{a}_t)
        -
        \left[
            r_t^{(h)}
            +
            \gamma^h
            Q_{\bar{\phi}}
            (s_{t+h},\mathbf{a}_{t+h})
        \right]
    \right)^2
\right].
\label{eq:qcfqlnstepcfp_critic_loss}
\end{equation}
\paragraph{QC-CFP.} QC naturally shares the same training approach as CFP—specifically, using only the actor network to approximate the velocity field. So its training objective is:
\begin{equation}
L_{\pi}(\theta)=\mathbb{E}_{\substack{
    x_0 \sim \mathcal{N}(0, I_{hd}), \\ 
    (x_1=a, s) \sim \mathcal{D}, \\ 
    t \sim \text{Unif}([0,1])
}}[\| v_\theta(t, s,x_t) - (x_1 - x_0) \|_2^2].
\end{equation}
For warm-up and online training, the objectives are:
\begin{equation}
L_{\pi}(\theta)=\mathbb{E}_{\substack{
    x_0 \sim \mathcal{N}(0, I_{hd}), \\ 
    (x_1=a, s) \sim \mathcal{D}, \\ 
    t \sim \text{Unif}([0,1])
}}[\| v_\theta(t, s,x_t) - (x_1 - x_0) \|_2^2],
\end{equation}
\begin{equation}
L_{Q}(\phi)
=
\mathbb{E}_{(s,\mathbf{a}_t,r,s')\sim\mathcal{D}}
\left[
    \left(
        Q_{\phi}(s_t,\mathbf{a_t})
        -
        \left[
            r_t^{(h)}
            +
            \gamma^h
            Q_{\bar{\phi}}
            \left(
                s_{t+h},
                \mathbf{a}_{t+h}^{\,i^*}
            \right)
        \right]
    \right)^2
\right],
\end{equation}
where $\mathbf{a^{\,i^*}_{t+h}}$ follows the Best-of-$N$ sampling.
\section{Experiment}
\subsection{Hyperparameters}
Unless otherwise specified, we use the same hyperparameters across all environments and algorithms. The common hyperparameters are summarized in Table~\ref{tab:common_hyperparameters}.

\begin{table}[htbp]
    \centering
    \caption{Common hyperparameters used in our experiments.}
    \label{tab:common_hyperparameters}
    \renewcommand{\arraystretch}{1.15}
    \begin{tabular}{@{}lc@{}}
        \toprule
        \multicolumn{1}{c}{\textbf{Parameter}}
        & \textbf{Value} \\
        \midrule
        Batch size (\(M\))                         & \(256\) \\
        Discount factor (\(\gamma\))               & \(0.99\) \\
        Optimizer                                   & Adam \\
        Learning rate                               & \(3\times10^{-4}\) \\
        Target network update rate (\(\tau\))       & \(5\times10^{-3}\) \\
        Critic ensemble size                        & \(2\) \\
        Number of flow integration steps (\(T\))
& \(4\) for QCFQL-nstep; \(10\) otherwise \\
        Action chunk length (\(h\))                 & \(5\) \\
        Number of offline training steps            & \(1\times10^{6}\) \\
        Number of online environment steps          & \(1\times10^{6}\) \\
        Network width                               & \(512\) \\
        Network depth                               & \(4\) hidden layers \\
        Number of best-of-\(N\) candidates (QC)     & \(32\) \\
        \bottomrule
    \end{tabular}
\end{table}

For specific hyperparameters of $\alpha$ value for backpropagation algorithms and $N$ for Best-of-$N$ algorithms, see Table \ref{tab:task_specific_hyperparameters}. For hyperparameters of warm-up steps, see Table \ref{tab:cfp_warmup_steps}.

\clearpage

\begin{table*}[p] 
\centering
\caption{Task-specific hyperparameters for all O2O and CFP variants.}
\label{tab:task_specific_hyperparameters}

\small
\renewcommand{\arraystretch}{1.08}

\begin{tabular*}{\textwidth}{
    @{\extracolsep{\fill}}
    ll
    cc
    cc
    cc
    cc
    @{}
}
\toprule
\multirow{3}{*}{Environment}
&
\multirow{3}{*}{Task}
&
\multicolumn{2}{c}{QC ($N$)}
&
\multicolumn{2}{c}{FQL ($\alpha$)}
&
\multicolumn{2}{c}{QCFQL-nstep ($\alpha$)}
&
\multicolumn{2}{c}{QCFQL ($\alpha$)}
\\
\cmidrule(lr){3-4}
\cmidrule(lr){5-6}
\cmidrule(lr){7-8}
\cmidrule(lr){9-10}
&
&
O2O & CFP
&
O2O & CFP
&
O2O & CFP
&
O2O & CFP
\\
\midrule

\multirow{5}{*}{Cube Double}
& Task 1 & 32 & 32 & 300 & 300 & 100 & 100 & 100 & 100 \\
& Task 2 & 32 & 32 & 300 & 300 & 100 & 100 & 100 & 100 \\
& Task 3 & 32 & 32 & 300 & 300 & 100 & 100 & 100 & 100 \\
& Task 4 & 32 & 32 & 300 & 300 & 100 & 100 & 100 & 100 \\
& Task 5 & 32 & 32 & 300 & 300 & 100 & 100 & 100 & 100 \\
\midrule

\multirow{5}{*}{Cube Triple}
& Task 1 & 32 & 32 & 300 & 300 & 100 & 100 & 100 & 100 \\
& Task 2 & 32 & 32 & 300 & 300 & 100 & 100 & 100 & 100 \\
& Task 3 & 32 & 32 & 300 & 300 & 100 & 100 & 100 & 100 \\
& Task 4 & 32 & 32 & 300 & 300 & 100 & 100 & 100 & 100 \\
& Task 5 & 32 & 32 & 300 & 300 & 100 & 100 & 100 & 100 \\
\midrule

\multirow{5}{*}{Cube Quadruple}
& Task 1 & 32 & 32 & 300 & 300 & 100 & 100 & 100 & 100 \\
& Task 2 & 32 & 32 & 300 & 300 & 100 & 100 & 100 & 100 \\
& Task 3 & 32 & 32 & 300 & 300 & 100 & 100 & 100 & 100 \\
& Task 4 & 32 & 32 & 300 & 300 & 100 & 100 & 100 & 100 \\
& Task 5 & 32 & 32 & 300 & 300 & 100 & 100 & 100 & 100 \\
\midrule

\multirow{5}{*}{Puzzle $4\times4$}
& Task 1 & 64 & 64 & 1000 & 1000 & 1000 & 1000 & 1000 & 1000 \\
& Task 2 & 64 & 64 & 1000 & 1000 & 1000 & 1000 & 1000 & 1000 \\
& Task 3 & 64 & 64 & 1000 & 1000 & 1000 & 100 & 1000 & 100 \\
& Task 4 & 64 & 64 & 1000 & 1000 & 1000 & 1000 & 1000 & 1000 \\
& Task 5 & 64 & 64 & 1000 & 1000 & 1000 & 1000 & 1000 & 1000 \\
\midrule

\multirow{5}{*}{Scene}
& Task 1 & 32 & 32 & 300 & 300 & 300 & 300 & 300 & 300 \\
& Task 2 & 32 & 32 & 300 & 300 & 300 & 300 & 300 & 300 \\
& Task 3 & 32 & 32 & 300 & 300 & 300 & 300 & 300 & 300 \\
& Task 4 & 32 & 32 & 300 & 300 & 300 & 300 & 300 & 300 \\
& Task 5 & 32 & 32 & 300 & 300 & 300 & 300 & 300 & 300 \\
\midrule
Lift & -- & 16 & 16 & 10000 & 10000 & 10000 & 10000 & 10000 & 10000 \\
Can & -- & 16 & 16 & 10000 & 10000 & 10000 & 10000 & 10000 & 10000 \\
Square & -- & 16 & 16 & 10000 & 10000 & 10000 & 10000 & 10000 & 10000 \\
\bottomrule
\end{tabular*}
\end{table*}

\clearpage
\begin{table*}[p] 
\centering
\caption{Task-specific critic warm-up steps for the CFP variants.}
\label{tab:cfp_warmup_steps}

\small
\renewcommand{\arraystretch}{1.08}

\begin{tabular*}{\textwidth}{
    @{\extracolsep{\fill}}
    llcccc
    @{}
}
\toprule
Environment
&
Task
&
QC-CFP
&
FQL-CFP
&
QCFQL-nstep-CFP
&
QCFQL-CFP
\\
\midrule

\multirow{5}{*}{Cube Double}
& Task 1 & 10000 & 10000 & 10000 & 10000 \\
& Task 2 & 10000 & 10000 & 10000 & 10000 \\
& Task 3 & 10000 & 10000 & 10000 & 10000 \\
& Task 4 & 10000 & 10000 & 10000 & 10000 \\
& Task 5 & 10000 & 10000 & 10000 & 10000 \\
\midrule

\multirow{5}{*}{Cube Triple}
& Task 1 & 10000 & 10000 & 10000 & 10000 \\
& Task 2 & 10000 & 10000 & 10000 & 10000 \\
& Task 3 & 10000 & 10000 & 10000 & 10000 \\
& Task 4 & 10000 & 10000 & 10000 & 10000 \\
& Task 5 & 10000 & 10000 & 0.1M & 10000 \\
\midrule

\multirow{5}{*}{Cube Quadruple}
& Task 1 & 0.2M & 0.2M & 0.2M & 0.2M \\
& Task 2 & 0.2M & 0.2M & 0.2M & 0.2M \\
& Task 3 & 0.2M & 0.2M & 0.2M & 0.2M \\
& Task 4 & 0.2M & 0.2M & 0.2M & 0.2M \\
& Task 5 & 0.2M & 0.2M & 0.2M & 0.2M \\
\midrule

\multirow{5}{*}{Puzzle $4\times4$}
& Task 1 & 10000 & 10000 & 10000 & 10000 \\
& Task 2 & 10000 & 10000 & 10000 & 10000 \\
& Task 3 & 50000 & 50000 & 50000 & 50000 \\
& Task 4 & 10000 & 50000 & 10000 & 10000 \\
& Task 5 & 10000 & 50000 & 10000 & 10000 \\
\midrule

\multirow{5}{*}{Scene}
& Task 1 & 10000 & 30000 & 10000 & 10000 \\
& Task 2 & 10000 & 30000 & 10000 & 10000 \\
& Task 3 & 10000 & 30000 & 10000 & 10000 \\
& Task 4 & 10000 & 30000 & 10000 & 10000 \\
& Task 5 & 10000 & 30000 & 10000 & 10000 \\
\midrule
Lift & Task 1 & 10000 & 5000 & 10000 & 10000 \\
Can & Task 1 & 10000 & 5000 & 10000 & 10000 \\
Square & Task 1 & 10000 & 5000 & 10000 & 5000 \\
\bottomrule
\end{tabular*}
\end{table*}

\clearpage
\subsection{Full Results}
\label{App:full_result}
\begin{figure*}[htbp]
    \centering
    \includegraphics[width=\textwidth]{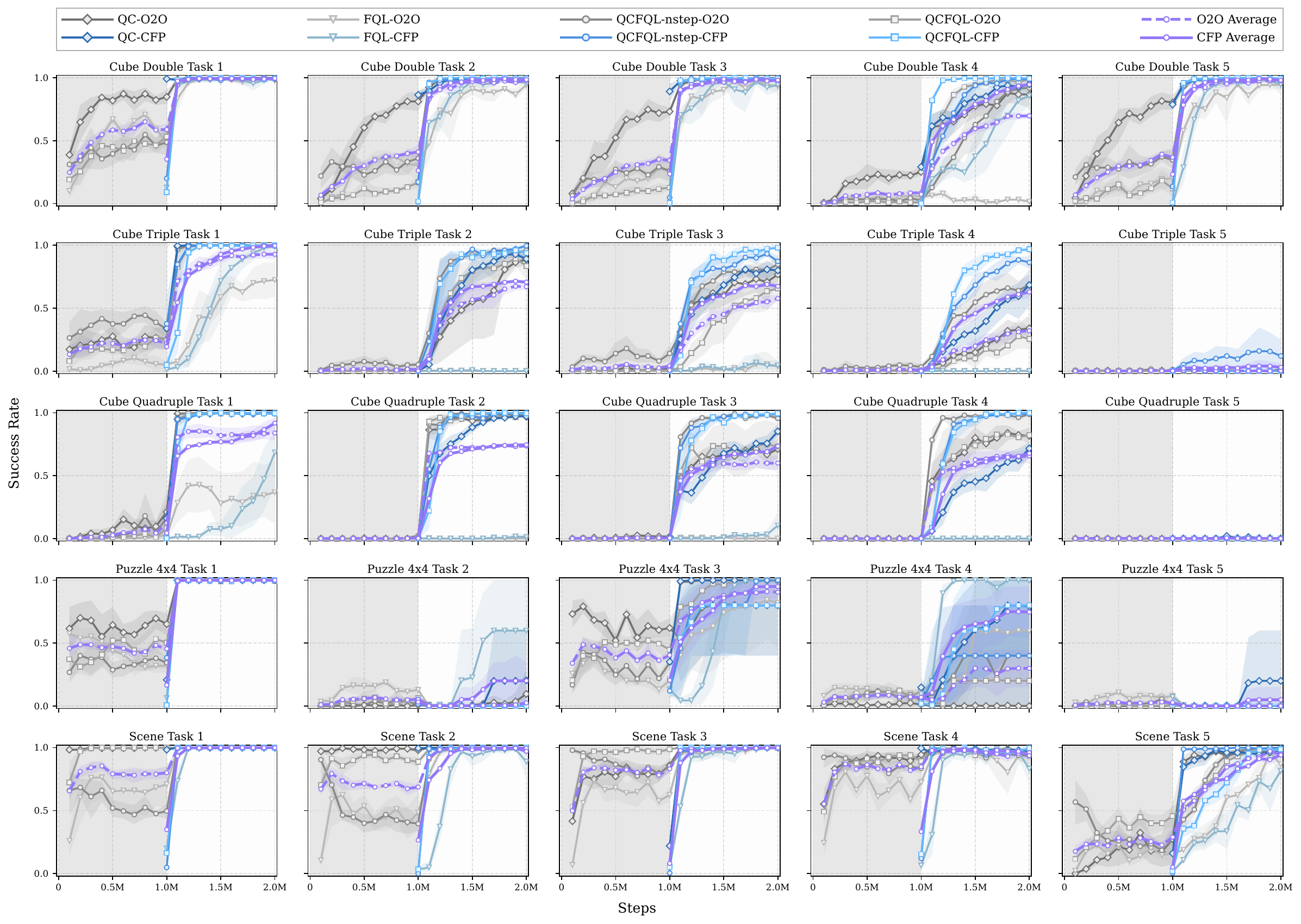}
    \caption{
        Full OGBench results by tasks. Curves and shaded regions represent the mean and 95\% confidence interval across five random seeds, respectively.
    }
    \label{fig:ogbench_full_results}
\end{figure*}

\paragraph{OGBench.} Figure \ref{fig:ogbench_full_results} shows the full results on OGBench.

\paragraph{Robomimic.} Figure \ref{fig:robomimic-main-results} shows the full results on Robomimic.

Table \ref{tab:task_level_full_results} shows the detailed performance comparison between O2O and CFP. Almost all CFP variants reach performance comparable to or even better than those of the O2O variants.
\begin{table*}[p]
\centering
\caption{Detailed  mean performance comparison (\%) by task.
All highlighted cells correspond to CFP.
\colorbox{commentblue}{\textcolor{white}{\textbf{Blue-shaded}}}
cells indicate that CFP significantly outperforms O2O;
\colorbox{algblue}{\textbf{light-blue-shaded}}
cells indicate minor variations or no significant difference; and
\colorbox{tablegray}{\textbf{gray-shaded}}
cells indicate that O2O significantly outperforms CFP.}
\label{tab:task_level_full_results}
\scriptsize
\setlength{\tabcolsep}{2.2pt} 
\renewcommand{\arraystretch}{1.18}
\begin{tabular*}{\textwidth}{@{\extracolsep{\fill}}ll cccccccc@{}}
\toprule
\multirow{2}{*}{Environment} & \multirow{2}{*}{Task} & \multicolumn{2}{c}{QC} & \multicolumn{2}{c}{FQL} & \multicolumn{2}{c}{QCFQL-nstep} & \multicolumn{2}{c}{QCFQL} \\
\cmidrule(lr){3-4} \cmidrule(lr){5-6} \cmidrule(lr){7-8} \cmidrule(lr){9-10} 
& & O2O & CFP & O2O & CFP & O2O & CFP & O2O & CFP \\
\midrule
\multirow{5}{*}{Cube Double} 
 & Task 1 & 84.8$\rightarrow$100 & \hlsmall{99.2$\rightarrow$100} & 48.8$\rightarrow$100 & \hlsmall{12.8$\rightarrow$100} & 48.8$\rightarrow$100 & \hlsmall{20.0$\rightarrow$100} & 53.2$\rightarrow$100 & \hlsmall{9.2$\rightarrow$100} \\
 & Task 2 & 81.2$\rightarrow$100 & \hlsmall{86.4$\rightarrow$100} & 30.4$\rightarrow$97.2 & \hlbig{0$\rightarrow$100} & 36.0$\rightarrow$100 & \hlsmall{17.2$\rightarrow$100} & 16.8$\rightarrow$100 & \hlsmall{1.6$\rightarrow$100} \\
 & Task 3 & 73.2$\rightarrow$100 & \hlsmall{89.2$\rightarrow$100} & 25.6$\rightarrow$98.8 & \hlsmall{0$\rightarrow$100} & 26.0$\rightarrow$100 & \hlsmall{4.8$\rightarrow$100} & 12.8$\rightarrow$100 & \hlsmall{0.4$\rightarrow$100} \\
 & Task 4 & 25.2$\rightarrow$90.0 & \hlbig{29.2$\rightarrow$96.4} & 5.6$\rightarrow$9.2 & \hlbig{0$\rightarrow$89.2} & 2.4$\rightarrow$94.4 & \hlsmall{0.4$\rightarrow$99.6} & 1.2$\rightarrow$100 & \hlsmall{0$\rightarrow$100} \\
 & Task 5 & 80.4$\rightarrow$100 & \hlsmall{78.8$\rightarrow$100} & 20.0$\rightarrow$97.6 & \hlbig{0$\rightarrow$100} & 34.8$\rightarrow$100 & \hlsmall{13.6$\rightarrow$100} & 12.0$\rightarrow$100 & \hlsmall{0.8$\rightarrow$100} \\
\midrule
\multirow{5}{*}{Cube Triple} 
 & Task 1 & 25.2$\rightarrow$100 & \hlsmall{34.0$\rightarrow$100} & 5.2$\rightarrow$78.8 & \hlbig{1.2$\rightarrow$99.2} & 33.2$\rightarrow$100 & \hlsmall{37.6$\rightarrow$100} & 24.4$\rightarrow$100 & \hlsmall{4.8$\rightarrow$100} \\
 & Task 2 & 0.8$\rightarrow$91.2 & \hlbig{0$\rightarrow$96.0} & 0$\rightarrow$1.2 & \hlsmall{0$\rightarrow$1.2} & 5.2$\rightarrow$97.6 & \hlsmall{0.4$\rightarrow$100} & 0$\rightarrow$92.0 & \hlbig{0$\rightarrow$98.4} \\
 & Task 3 & 2.0$\rightarrow$80.4 & \hlsmall{2.8$\rightarrow$84.0} & 0.4$\rightarrow$8.4 & \hlbig{0$\rightarrow$13.2} & 14.0$\rightarrow$87.6 & \hlbig{0$\rightarrow$95.6} & 0$\rightarrow$68.8 & \hlbig{0$\rightarrow$99.6} \\
 & Task 4 & 0$\rightarrow$38.0 & \hlbig{0$\rightarrow$68.4} & 0$\rightarrow$0.4 & \hlsmall{0$\rightarrow$0.4} & 3.2$\rightarrow$75.6 & \hlbig{0$\rightarrow$91.6} & 0$\rightarrow$30.8 & \hlbig{0$\rightarrow$98.4} \\
 & Task 5 & 0$\rightarrow$2.4 & \hlsmall{0$\rightarrow$1.6} & 0$\rightarrow$0 & \hlsmall{0$\rightarrow$0} & 1.2$\rightarrow$2.4 & \hlbig{0.4$\rightarrow$25.6} & 0$\rightarrow$0 & \hlbig{0$\rightarrow$4.0} \\
\midrule
\multirow{5}{*}{Cube Quadruple} 
 & Task 1 & 21.2$\rightarrow$100 & \hlsmall{16.0$\rightarrow$100} & 2.4$\rightarrow$48.8 & \hlbig{0$\rightarrow$68.4} & 16.8$\rightarrow$100 & \hlsmall{1.6$\rightarrow$100} & 8.4$\rightarrow$100 & \hlsmall{0$\rightarrow$100} \\
 & Task 2 & 0$\rightarrow$99.6 & \hlsmall{0.4$\rightarrow$98.8} & 0$\rightarrow$0.4 & \hlbig{0$\rightarrow$2.8} & 2.4$\rightarrow$100 & \hlsmall{0$\rightarrow$100} & 0$\rightarrow$99.6 & \hlsmall{0$\rightarrow$100} \\
 & Task 3 & 1.6$\rightarrow$76.8 & \hlbig{1.2$\rightarrow$86.4} & 0$\rightarrow$2.0 & \hlbig{0$\rightarrow$10.8} & 0$\rightarrow$100 & \hlsmall{0.4$\rightarrow$100} & 0$\rightarrow$80.0 & \hlbig{0$\rightarrow$100} \\
 & Task 4 & 0$\rightarrow$89.6 & \hlgray{0$\rightarrow$71.6} & 0$\rightarrow$0 & \hlsmall{0$\rightarrow$0} & 0$\rightarrow$99.6 & \hlsmall{0$\rightarrow$100} & 0$\rightarrow$86.8 & \hlbig{0$\rightarrow$100} \\
 & Task 5 & 0$\rightarrow$0 & \hlbig{0$\rightarrow$2.4} & 0$\rightarrow$0 & \hlsmall{0$\rightarrow$0} & 0$\rightarrow$0 & \hlsmall{0$\rightarrow$0} & 0$\rightarrow$0 & \hlsmall{0$\rightarrow$0} \\
\midrule
\multirow{5}{*}{Puzzle 4x4} 
 & Task 1 & 65.2$\rightarrow$100 & \hlsmall{20.8$\rightarrow$100} & 32.0$\rightarrow$100 & \hlsmall{6.4$\rightarrow$100} & 34.4$\rightarrow$100 & \hlsmall{38.4$\rightarrow$100} & 50.4$\rightarrow$100 & \hlsmall{0.8$\rightarrow$100} \\
 & Task 2 & 0.8$\rightarrow$10.0 & \hlbig{3.2$\rightarrow$22.8} & 12.8$\rightarrow$12.8 & \hlbig{2.8$\rightarrow$62.0} & 0.8$\rightarrow$0.8 & \hlsmall{1.6$\rightarrow$1.6} & 6.4$\rightarrow$6.8 & \hlsmall{5.2$\rightarrow$5.2} \\
 & Task 3 & 62.0$\rightarrow$90.0 & \hlbig{35.2$\rightarrow$100} & 20.4$\rightarrow$84.8 & \hlbig{13.2$\rightarrow$100} & 34.4$\rightarrow$100 & \hlsmall{12.0$\rightarrow$100} & 44.8$\rightarrow$100 & \hlgray{21.6$\rightarrow$82.0} \\
 & Task 4 & 1.2$\rightarrow$1.2 & \hlbig{14.8$\rightarrow$85.2} & 9.6$\rightarrow$64.4 & \hlbig{9.2$\rightarrow$100} & 4.0$\rightarrow$42.0 & \hlsmall{2.0$\rightarrow$40.4} & 12.0$\rightarrow$29.2 & \hlbig{2.0$\rightarrow$80.4} \\
 & Task 5 & 0.4$\rightarrow$0.4 & \hlbig{2.0$\rightarrow$22.4} & 8.0$\rightarrow$8.0 & \hlsmall{7.2$\rightarrow$7.2} & 0.4$\rightarrow$0.4 & \hlsmall{1.2$\rightarrow$1.2} & 0.4$\rightarrow$0.4 & \hlsmall{1.2$\rightarrow$1.2} \\
\midrule
\multirow{5}{*}{Scene} 
 & Task 1 & 100$\rightarrow$100 & \hlsmall{98.4$\rightarrow$100} & 70.8$\rightarrow$100 & \hlsmall{20.0$\rightarrow$100} & 49.6$\rightarrow$100 & \hlsmall{4.8$\rightarrow$100} & 100$\rightarrow$100 & \hlsmall{16.4$\rightarrow$100} \\
 & Task 2 & 98.0$\rightarrow$100 & \hlsmall{99.6$\rightarrow$100} & 48.4$\rightarrow$100 & \hlsmall{3.2$\rightarrow$100} & 39.6$\rightarrow$100 & \hlsmall{0$\rightarrow$100} & 88.8$\rightarrow$100 & \hlsmall{3.2$\rightarrow$100} \\
 & Task 3 & 86.0$\rightarrow$100 & \hlsmall{22.0$\rightarrow$100} & 62.4$\rightarrow$100 & \hlsmall{4.4$\rightarrow$100} & 85.6$\rightarrow$100 & \hlsmall{0.4$\rightarrow$100} & 98.8$\rightarrow$100 & \hlsmall{5.2$\rightarrow$100} \\
 & Task 4 & 94.4$\rightarrow$100 & \hlsmall{99.6$\rightarrow$100} & 72.4$\rightarrow$100 & \hlsmall{6.4$\rightarrow$100} & 83.6$\rightarrow$100 & \hlsmall{12.0$\rightarrow$100} & 89.2$\rightarrow$100 & \hlsmall{15.2$\rightarrow$100} \\
 & Task 5 & 27.2$\rightarrow$99.6 & \hlsmall{16.0$\rightarrow$99.6} & 15.6$\rightarrow$96.0 & \hlgray{1.6$\rightarrow$85.2} & 27.2$\rightarrow$100 & \hlsmall{0.4$\rightarrow$100} & 45.6$\rightarrow$100 & \hlsmall{2.0$\rightarrow$100} \\

\midrule
Lift & --
& 92.8$\rightarrow$100 & \hlsmall{91.6$\rightarrow$100}
& 79.2$\rightarrow$98.8 & \hlsmall{81.6$\rightarrow$98.4}
& 96.4$\rightarrow$100 & \hlsmall{96.0$\rightarrow$100}
& 98.0$\rightarrow$100 & \hlsmall{95.6$\rightarrow$100} \\
Can & --
& 80.8$\rightarrow$97.2 & \hlsmall{80.8$\rightarrow$96.8}
& 26.4$\rightarrow$77.6 & \hlsmall{35.6$\rightarrow$76.4}
& 82.4$\rightarrow$99.6 & \hlsmall{81.6$\rightarrow$99.6}
& 83.6$\rightarrow$98.8 & \hlsmall{90.0$\rightarrow$97.6} \\
Square & --
& 38.0$\rightarrow$94.4 & \hlsmall{37.2$\rightarrow$94.0}
& 6.0$\rightarrow$23.2 & \hlgray{4.4$\rightarrow$12.4}
& 30.8$\rightarrow$73.2 & \hlbig{30.4$\rightarrow$82.4}
& 32.4$\rightarrow$82.8 & \hlsmall{27.2$\rightarrow$80.4} \\
\bottomrule
\end{tabular*}
\end{table*}
\clearpage
\subsection{Ablation Study}
\label{App:ablation study}
\begin{figure}[H]
    \centering
    \includegraphics[
        width=\textwidth
    ]{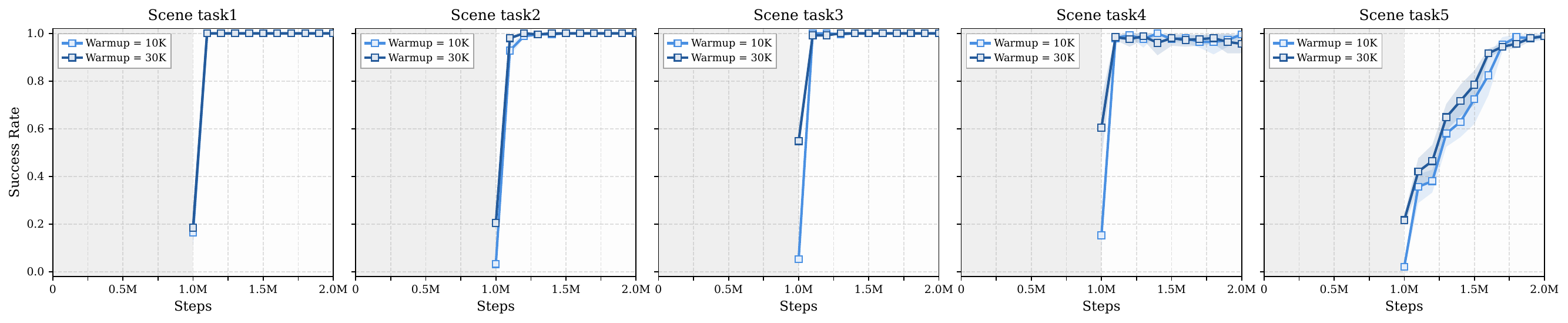}
    \caption{
        Ablation of the number of steps during warm-up on OGBench. The applied algorithm is QCFQL-CFP. Curves and shaded regions represent the mean and 95\%{} confidence
        interval across five random seeds, respectively.
    }
    \label{fig:ablation_steps_appendix}
\end{figure}

We present more results of the ablation study here. Figure \ref{fig:ablation_steps_appendix} further confirms our conclusion that CFP is insensitive to the steps of warm-up stage. 

\begin{figure}[htbp]
    \centering
    \includegraphics[
        width=\textwidth
    ]{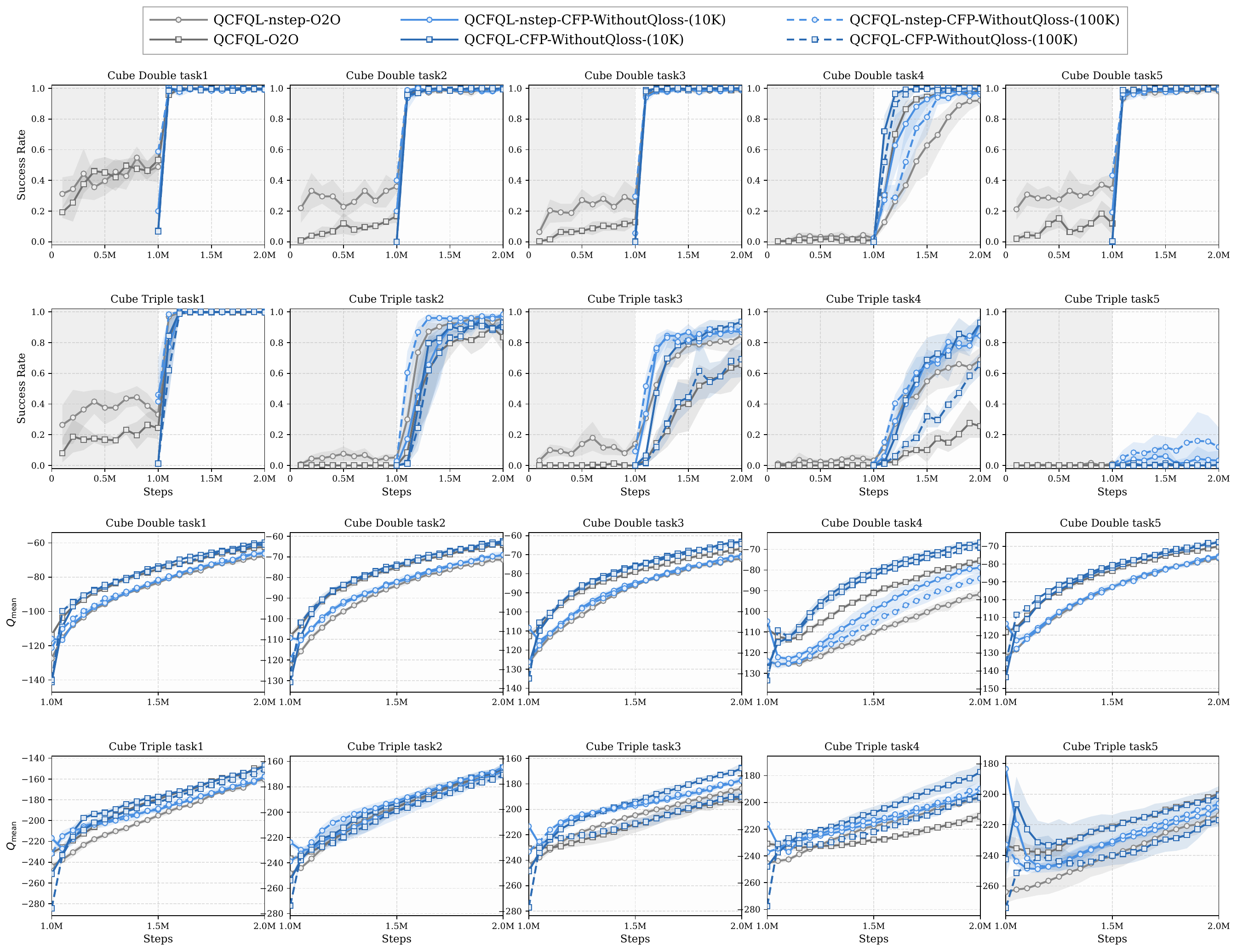}
    \caption{
        Ablation study of the sensitivity of CFP (without Q-loss) to warm-up steps.
    }
    \label{fig:withoutqloss_warmup_steps}
\end{figure}

We also investigate CFP's sensitivity to warm-up steps in the absence of Q-loss during the warm-up phase. Figure \ref{fig:withoutqloss_warmup_steps} shows that CFP without Q-loss during warm-up stage is more sensitive to the length of warm-up steps. Overall, the longer the warm-up period, the worse the performance of online fine-tuning of CFP. This result is significantly different from the conclusion we make in section \ref{sec:warm-up steps}. We examined the corresponding mean Q-values and find that the Q mean assigned by the critic decreased as the warm-up length increased. We attribute this phenomenon to the interplay within the actor-critic architecture. We have demonstrated that a fresh critic can provide guidance even during the warm-up phase, enabling the actor to output higher-value actions; conversely, since the target action in the critic's update formula is provided by the actor, higher-quality actions from the actor result in higher action ratings from the critic. Without the inclusion of Q-loss, the actor remains confined to the low-quality actions present in the dataset, producing progressively lower critic estimates of state-action pairs. Our research further confirms that offline datasets—characterized by low environmental coverage and limited trajectory success rates—exacerbate the critic's pessimism, thereby hindering generalization during the online phase.

\clearpage
\section{Toy Example}
\label{App:toy_exp}
In this section, we introduce the toy example we implement in Section \ref{sec: why initialized critic helps with online Fine-tuning} in detail.

We designed an MDP table and simulated the complete O2O phase as well as the CFP implementation during the actual training process. In our experiments, we rigorously controlled the variables associated with O2O and CFP, ensuring that the sole difference between the two was whether the critic was initialized during the online phase.
\subsection{Environment}

The environment is a finite MDP with $16$ states.  For visualization, the
states are placed on a $4\times4$ canvas,
\begin{equation}
\begin{matrix}
0 & 1 & 2 & 3 \\
4 & 5 & 6 & 7 \\
8 & 9 & 10 & 11 \\
12 & 13 & 14 & 15,
\end{matrix}
\end{equation}
but these coordinates do not define the transition dynamics.  The dynamics
are instead specified by an irregular directed graph.  An action at state
$s$ selects one of its valid destination states $s'\in\mathcal N^+(s)$;
therefore the action-value function is represented as a matrix $Q(s,s')$.
The graph includes both local edges and nonlocal directed shortcuts, and the
existence of $s\rightarrow s'$ does not imply the existence of
$s'\rightarrow s$. Table \ref{tab:toy_transition_graph} provides specific information. The figure in toy example \ref{box:toy-example} also demonstrates the same dynamics in a visual way.

\begin{table}[H]
\centering
\caption{Possible transitions of the sixteen-state directed MDP.}
\label{tab:toy_transition_graph}

\setlength{\tabcolsep}{20pt}
\begin{tabular}{c l @\qquad c l}
\toprule
$s$ & $\mathcal N^+(s)$ & $s$ & $\mathcal N^+(s)$ \\
\midrule
0  & $\varnothing$       & 8  & $\{1,4,9,12,14\}$ \\
1  & $\{0,2,5,9\}$       & 9  & $\{2,5,8,10,15\}$ \\
2  & $\{1,6,8,11\}$      & 10 & $\{3,6,9,11,14\}$ \\
3  & $\{2,7,10,14\}$     & 11 & $\{4,7,10,15\}$ \\
4  & $\{0,5,8,11\}$      & 12 & $\{0,5,8,13\}$ \\
5  & $\{1,4,6,10,13\}$   & 13 & $\{4,7,9,12,14\}$ \\
6  & $\{2,5,7,9,14\}$    & 14 & $\{3,6,10,13,15\}$ \\
7  & $\{3,6,11,13\}$     & 15 & $\varnothing$ \\
\bottomrule
\end{tabular}
\end{table}

State 0 is a bad terminal with reward $-8$, whereas state 15 is a good terminal with reward $+12$. Every other transition has a base reward of $-0.05$. Entering one of the hazard states
$\mathcal H=\{5,10,12\}$ incurs an additional penalty of $-1.25$.
Several directed shortcuts also have edge-specific costs, listed in
Table \ref{tab:toy_shortcut_costs}.  Thus, for a nonterminal destination,
\begin{equation}
r(s,s')=-0.05-1.25\,\mathbb I[s'\in\mathcal H]+c(s,s'),
\end{equation}
while transitions into states 0 and 15 receive their terminal rewards. The discount factor is $\gamma=0.97$.
\begin{table}[H]
\centering
\caption{Edge-specific costs $c(s,s')$.}
\label{tab:toy_shortcut_costs}
\setlength{\tabcolsep}{20pt}
\begin{tabular}{c @\quad r  c @\quad r  c @\quad r}
\toprule
Edge & Cost & Edge & Cost & Edge & Cost \\
\midrule
$1\to9$   & $-0.70$ & $2\to11$ & $+0.15$ & $3\to14$ & $-0.40$ \\
$4\to11$  & $-0.20$ & $5\to13$ & $-0.60$ & $6\to14$ & $+0.10$ \\
$7\to13$  & $-0.30$ & $8\to1$  & $-0.50$ & $9\to2$  & $-0.40$ \\
$10\to3$  & $-0.20$ & $11\to4$ & $-0.60$ & $12\to5$ & $-0.25$ \\
$13\to4$  & $-0.35$ & $14\to3$ & $-0.45$ & & \\
\bottomrule
\end{tabular}
\end{table}
To simulate the inability of the offline dataset to provide optimal or suboptimal action coverage in complex environments, we designed the offline dataset to tend to reach the bad terminal. Specifically, for every nonterminal state, we compute the shortest directed distance from each valid successor to state 0. A probability mass of $0.985$ is distributed among the successors with minimum distance to state 0, while the remaining mass is
distributed uniformly over all valid successors.  This policy generates an offline dataset whose dominant trajectories terminate at state 0. 

Furthermore, to simulate a policy that more closely resembles the one operating in the real world environment---rather than the dataset policy---after the actor has undergone fine-tuning during the online phase, we implemented an online sampling strategy. Specifically, the prescribed online behavior is an $\epsilon$-greedy policy with respect to the exact optimal Q-function of the finite MDP, with $\epsilon=0.05$. It therefore predominantly follows high-value routes toward the positive terminal state 15. This policy is used only to generate state-action samples that model the rapid actor-distribution shift observed after a small number of online Q-loss updates.

For clarity, the implementation uses three distinct datasets:
\begin{enumerate}
    \item \textbf{Offline dataset} $\mathcal D_{\mathrm{off}}$: generated by
    the policy attracted to terminal state 0; used to fit the offline flow and the offline critic.
    \item \textbf{Post-shift actor dataset} $\mathcal D_{\mathrm{actor}}$:
    generated by the prescribed near-optimal online behavior.
    \item \textbf{Online critic replay} $\mathcal D_{\mathrm{replay}}$:
    generated by
    \begin{equation}
    \pi_{\mathrm{replay}}
    =0.70\,\pi_{\mathrm{near\text{-}optimal}}
    +0.30\,\pi_{\mathrm{uniform}},
    \end{equation}
    where $\pi_{\mathrm{uniform}}$ is uniform over valid outgoing edges.
\end{enumerate}
By decoupling the online actor and the online critic (regardless of whether offline pre-training is employed), we ensure that the actor remains high-performing and reliable during the online phase; this prevents low-quality offline data from impairing the actor's exploration capabilities—and, consequently, the training of the online critic. Similarly, by incorporating a uniform policy into the online critic's training data, we simulate the presence of low-quality data that might otherwise affect the critic's training.

Each dataset contains 35,000 episodes.  An episode begins from a uniformly sampled nonterminal state and ends at a terminal state or after 80 transitions. Truncated transitions are treated as terminal in the TD mask.
\subsection{Actor}

Each discrete destination state is embedded using its normalized two-dimensional
canvas coordinate in $[-1,1]^2$.  Given a state $s$, Gaussian noise $x_0$, and
flow time $t$, the flow network predicts a velocity field.  For a dataset
action $x_1$, Flow Matching minimizes
\begin{align}
x_t &= (1-t)x_0+t x_1,\\
L_{\mathrm{Flow}}
&=\mathbb E\!\left[
\left\|v_\theta(s,x_t,t)-(x_1-x_0)\right\|_2^2
\right].
\end{align}
The flow network contains two hidden layers of width 128.  At inference time,
we use 10 Euler integration steps, clip the continuous action to $[-1,1]^2$, and project it to the closest valid destination of the current state.

We first train an offline flow $\hat\pi_{\mathrm{off}}$ on
$\mathcal D_{\mathrm{off}}$.  Starting from this fitted flow, we then perform
500 additional flow updates on $\mathcal D_{\mathrm{actor}}$ and freeze the
resulting post-shift policy $\hat\pi_{\mathrm{on}}$. 
\subsection{Critic}

The critic concatenates a one-hot state vector with
the two-dimensional destination embedding, applies a single hidden layer of
width 48 with a $\tanh$ activation, and outputs one scalar Q-value. The critic minimizes
\begin{equation}
L_Q(\phi)=\mathbb E_{(s,a,r,s')\sim\mathcal D, a'\sim\hat\pi(\cdot\mid s')}
\left[
\left(
Q_\phi(s,a)-
\left[r+\gamma  Q_{\bar\phi}(s',a')\right]
\right)^2
\right]
\end{equation}
The target parameters are updated with $\tau=0.02$.
\subsection{Implementation}

The complete training pipeline is as follows.
\begin{enumerate}
    \item Initialize one flow network and fit it to
    $\mathcal D_{\mathrm{off}}$, obtaining $\hat\pi_{\mathrm{off}}$.
    \item Initialize a small critic once and save its random parameters
    $\phi_{\mathrm{init}}$.
    \item Train that critic for 500 TD updates on
    $\mathcal D_{\mathrm{off}}$ using $\hat\pi_{\mathrm{off}}$ in the TD
    target, obtaining $\phi_{\mathrm{off}}$.
    \item Adapt the flow policy on $\mathcal D_{\mathrm{actor}}$ and freeze the resulting post-shift policy $\hat\pi_{\mathrm{on}}$.
    \item Draw one common sequence of online-replay minibatches and one common
    sequence of Gaussian policy-noise keys.
    \item Run O2O from $\phi_{\mathrm{off}}$ and CFP from the saved
    $\phi_{\mathrm{init}}$. 
    Then perform 500 TD updates on the same replay using the same frozen $\hat\pi_{\mathrm{on}}$, minibatches, and policy noises.
\end{enumerate}
\subsection{Hyperparameters}
Table \ref{tab:toy_hyperparameters} summarizes the hyperparameters used in our toy example.

\begin{table*}[htbp]
\centering
\caption{Hyperparameters of the toy example.}
\label{tab:toy_hyperparameters}
\begin{tabular}{l c @\qquad l c}
\toprule
Hyperparameter & Value & Hyperparameter & Value \\
\midrule
Batch size & 256 & Discount factor $\gamma$ & 0.97 \\
Offline episodes & 35,000 & Online actor episodes & 35,000 \\
Online replay episodes & 35,000 & Maximum episode length & 80 \\
Offline-policy greed & 0.985 & Online-policy $\epsilon$ & 0.05 \\
Uniform replay mixture & 0.30 & Flow steps & 10 \\
Actor hidden widths & [$128,128$] &  Actor learning rate & $3\times10^{-4}$ \\
Critic hidden width & 48 & Critic learning rate & $10^{-3}$ \\
Offline updates & 500 & Online updates & 500 \\
Target update $\tau$ & 0.02 & RMSE recording interval & 10 \\

\bottomrule
\end{tabular}
\end{table*}

\subsection{Results and Interpretation}
\begin{figure}[htbp]
    \centering
    \includegraphics[
        width=\textwidth
    ]{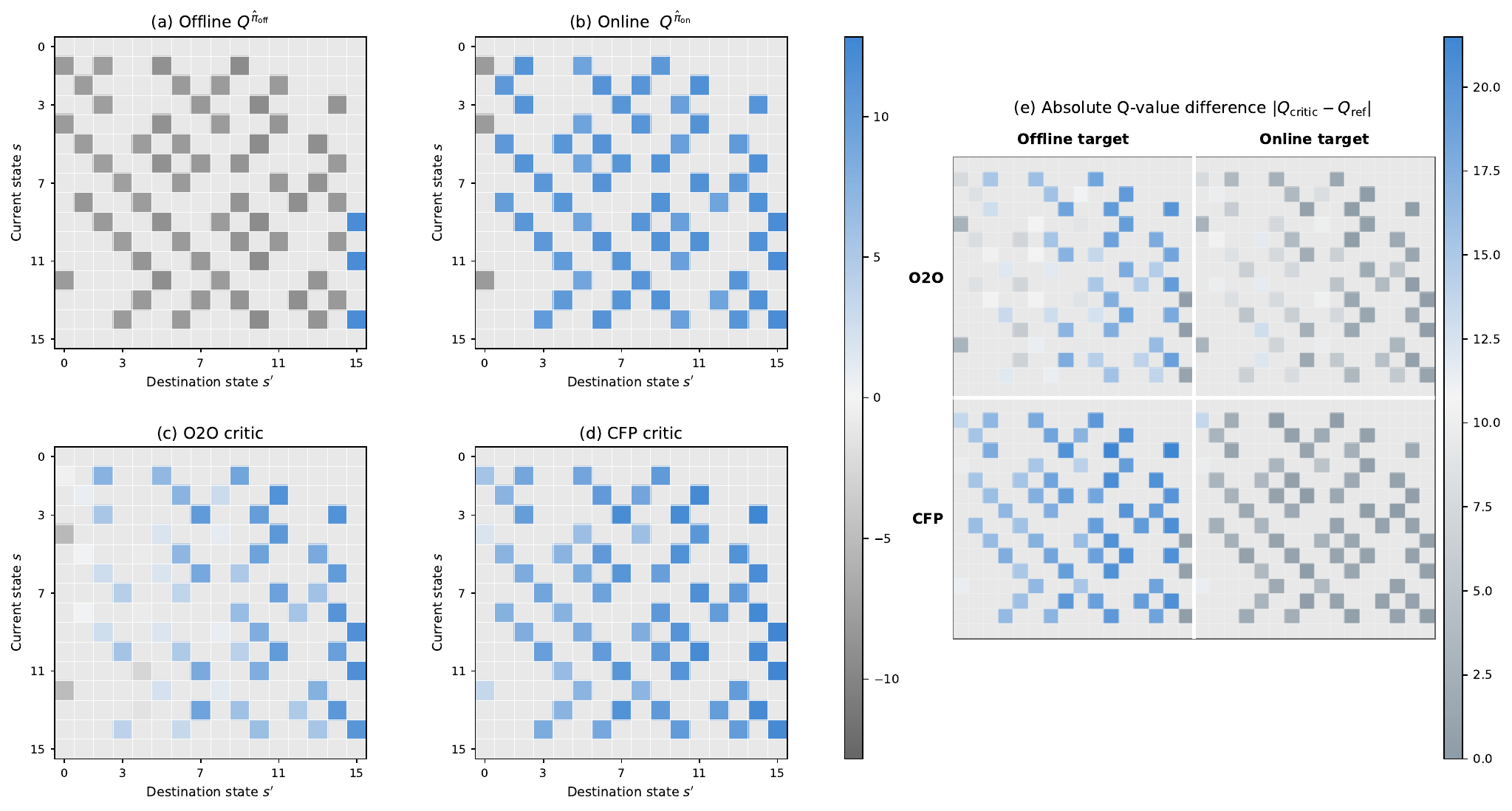}
    \caption{
        Results of toy example.
    }
    \label{fig:toy_example_result}
\end{figure}

We calculated the target Q-values for the offline and online phases based on offline and online strategies, respectively. We also evaluate the Q-values produced by the O2O and CFP critics. The left side of the figure \ref{fig:toy_example_result} displays the theoretical target Q-values for the offline and online settings, as well as the Q-values output by the O2O and CFP critics, respectively. The right side shows heatmaps representing the absolute differences between pairs of the plots on the left; areas appearing more gray indicate greater similarity, while darker blue areas indicate larger differences. The O2O critic output is significantly closer to the target value of the offline dataset than the CFP critic output. Conversely, the CFP critic output is closer to the online target Q-value.
\end{document}